\documentclass[]{fairmeta}
\usepackage{wrapfig}
\usepackage{array}
\usepackage{tabularx}
\usepackage{textcomp}
\usepackage{stfloats}
\usepackage{url}
\usepackage{verbatim}
\usepackage{graphicx}
\usepackage{titlesec}
\usepackage{tocloft}
\usepackage{adjustbox}
\usepackage{multirow}
\usepackage{pifont}
\usepackage{tikz}
\usepackage{comment}
\usepackage{amsmath,amssymb} 
\usepackage{colortbl}  
\usepackage[dvipsnames]{xcolor}         
\usepackage{booktabs} 
\usepackage{hyperref}
\usepackage{graphicx}    
\usepackage{subcaption} 
\usepackage{booktabs}
\usepackage{amsmath} 
\usepackage{multirow} 
\usepackage{booktabs} 
\usepackage{subcaption} 
\usepackage{todonotes}
\usepackage{natbib}
\usepackage{makecell}
\usepackage{tcolorbox,enumitem,ragged2e}
\tcbuselibrary{breakable}
\tcbset{
  promptbox/.style={
    colback=black!2, colframe=black!20, boxrule=0.3pt,
    arc=1mm, left=6pt, right=6pt, top=6pt, bottom=6pt,
    breakable}
}

\RequirePackage{xspace}
\makeatletter
\DeclareRobustCommand\onedot{\futurelet\@let@token\@onedot}
\def\@onedot{\ifx\@let@token.\else.\null\fi\xspace}

\makeatother

\definecolor{adptorange}{RGB}{248, 205, 172}
\definecolor{cmpblue}{RGB}{189, 215, 238}
\definecolor{cmpblue}{RGB}{189, 215, 238}

\definecolor{our_red}{RGB}{232,157,160}
\definecolor{our_blue}{RGB}{136,206,230}
\definecolor{our_orange}{RGB}{246,200,168}
\definecolor{our_green}{RGB}{178,211,164}

\definecolor{attn_code0}{RGB}{247,215,200}
\definecolor{attn_code1}{RGB}{238,169,139}
\definecolor{mlp_code0}{RGB}{204,201,221}
\definecolor{mlp_code1}{RGB}{102,95,153}

\definecolor{dark_green}{rgb}{0, 0.5, 0}
\definecolor{dark_red}{rgb}{0.8, 0.2, 0.2}
\definecolor{soft_red}{rgb}{1.0, 0.4, 0.4}
\definecolor{light_blue}{rgb}{0.2, 0.5, 1.0}

\usepackage{algpseudocode}
\usepackage{algorithm}

\definecolor{token_blue}{RGB}{84, 120, 140}

\usepackage{pifont}       
\usepackage{bbding}       
\usepackage{fontawesome}
\usepackage{xspace}
\newcommand{\xmark}{\ding{55}}
\newcommand{\cmark}{\ding{51}}
\usepackage{float}
\usepackage{placeins}
\usepackage{siunitx}        
\usepackage{microtype}      
\usepackage{algorithm}     
\usepackage{algorithmicx}  
\usepackage{algpseudocode} 
\usepackage{cleveref}      
\usepackage{enumitem}
\usepackage[inkscapelatex=false]{svg}
\usepackage{animate}
\usepackage{etoc}          

\definecolor{darkgreen}{rgb}{0.15, 0.75, 0.15}
\definecolor{cvprblue}{rgb}{0.21,0.49,0.74}
\definecolor{lightblue}{rgb}{0.90, 0.95, 0.99}

\algrenewcommand\algorithmicrequire{\textbf{Input:}}
\algrenewcommand\algorithmicensure{\textbf{Output:}}

\title{
  WorldOlympiad: Can Your World Model Survive a Triathlon?
}
\author[1*]{Yuke Zhao}
\author[3*]{Wangbo Zhao}
\author[1*]{Weijie Wang}
\author[2*\dag]{Zeyu Zhang}
\author[3]{Dakai An}
\author[4]{Akide Liu}
\author[5]{Yinghao Yu}
\author[2\ddag]{Jiasheng Tang}
\author[2]{Fan Wang}
\author[3]{Wei Wang}
\author[1\ddag]{Bohan Zhuang}

\affiliation[1]{Zhejiang University}
\affiliation[2]{DAMO Academy, Alibaba Group}
\affiliation[3]{The Hong Kong University of Science and Technology}
\affiliation[4]{Monash University}
\affiliation[5]{TRE, Alibaba Group}
\contribution[*]{Equal contribution.}
\contribution[\dag]{Project lead.}
\contribution[\ddag]{Corresponding authors.}

\abstract{
We introduce WorldOlympiad, a benchmark for diagnosing video-based world models across physical faithfulness, geometric consistency, and interaction fidelity. While existing benchmarks often focus on visual quality, semantic alignment, or short-term temporal coherence, they provide limited insight into whether generated videos obey physical rules, preserve coherent 3D structure, and sustain controllable interactions over long horizons. To address this gap, WorldOlympiad decomposes world-model evaluation into three complementary dimensions. The physical track uses object segmentation and MLLM-as-judge to assess whether generated videos follow interpretable rules in mechanics, thermal phenomena, and material properties. The geometry track reconstructs generated videos with Gaussian splatting and evaluates structural consistency, cross-view coherence, and camera-trajectory alignment. The interaction track assesses whether generated rollouts follow complex action prompts and maintain smooth, coherent transitions across consecutive video chunks. WorldOlympiad further covers three major downstream scenarios, including gaming, robotics, and general real-world videos, capturing diverse challenges from interactive control and embodied manipulation to open-domain motion and camera dynamics. Together, these tracks and scenarios form a scalable and interpretable evaluation suite that exposes failure modes beyond generic video quality. Experiments on state-of-the-art models reveal substantial gaps in physical reasoning, 3D consistency, and long-horizon interaction, underscoring the need for more structured evaluation protocols for generative world models.

}

\date{\today} 
\metadata[Project website]{\url{https://alibaba-damo-academy.github.io/WorldOlympiad}}
\metadata[Code]{\url{https://github.com/alibaba-damo-academy/WorldOlympiad}}
\metadata[Correspondence]{\email{jiasheng.tjs@alibaba-inc.com}, \email{bohan.zhuang@gmail.com}}

\begin{document}
\thispagestyle{firstheader}
\maketitle

\section{Introduction}
Recent advances in video generation~\citep{wan2025wan,team2025longcat,kong2024hunyuanvideo,yang2024cogvideox,blattmann2023stable,brooks2024video} have expanded its scope from passive content creation to video-based world modeling. A video-based world model is expected to predict future visual states from historical observations and control signals, which is crucial for game simulation~\citep{che2025gamegen,zhang2025matrix,he2025matrix}, robotic policy development~\citep{agarwal2025cosmos,ali2025world,chi2025wow}, and real-world scene generation~\citep{mao2025yume,team2026advancing,wu2026infinite,sun2025worldplay,yang2025longlive}. In these applications, high visual fidelity alone is insufficient: models must preserve state continuity, respect physical and geometric constraints, respond to user actions, and maintain plausible dynamics over long generation horizons. 
Together, these requirements make developing a comprehensive framework for evaluating video-based world models across multiple capability dimensions a key challenge.


To address this challenge, a straightforward solution is to reuse evaluation approaches originally designed for traditional video generation models, such as VBench~\citep{huang2024vbench}, VBench-2.0~\citep{zheng2025vbench}, and CLIP-based metrics~\citep{yang2025longlive,huang2025self,liu2025rolling}.  However, these approaches primarily measure perceptual quality or text--video alignment on short videos and fail to fully capture core world-modeling properties. Although the the subsequent VBench++~\citep{huang2025vbench++} extends  evaluation to long-video generation, it still focuses largely on visual appearance and temporal smoothness, leaving key world-modeling capabilities underexplored. In particular, existing approaches pay limited attention to whether generated videos can consistently adhere to physical laws, preserve coherent 3D structure, and support controllable interactions over long horizons.

Moreover, recent benchmarks specifically designed for video-based world models often focus on a single downstream domain, such as gaming~\citep{ye2026mind} or robotics~\citep{shang2026worldarena,deng2026rethinking,liu2026rise}, making it
difficult to compare models under a unified protocol across gaming, robotics,
and general real-world scenarios. As a result, current benchmarks still cannot fully answer a central question: Can existing video-based world models reliably simulate world dynamics across multiple domains, over long horizons, and in interactive settings?


Building on prior efforts, we identify two key challenges in evaluating video-based world models: developing capability-oriented metrics and ensuring task diversity. Evaluation should span heterogeneous domains, while the metrics should assess whether generated videos maintain long-horizon consistency, obey physical laws, preserve 3D geometry, and respond faithfully to control signals. Although existing benchmarks partially address these challenges, they do not comprehensively cover both aspects, as summarized in Table~\ref{tab:benchmark_comparison}.

To this end, we introduce \textbf{WorldOlympiad}, a unified benchmark for evaluating video-based world models across gaming, robotics, and real-world scenarios. WorldOlympiad is constructed from 1,000 high-quality long videos that cover diverse downstream requirements, including interactive control in games, embodied manipulation in robotics, and open-domain motion and camera dynamics in real-world scenes. Each video is processed into structured evaluation instances with scenario-specific prompts and chunk-level temporal descriptions, enabling controlled long-video generation and fine-grained diagnosis. Built on this benchmark, we evaluate generated videos from three complementary perspectives: physical faithfulness, geometric consistency, and interaction fidelity. With 8 representative long-video generation pipelines, WorldOlympiad reveals systematic limitations in current models and provides diagnostic evidence for developing more reliable video-based world models.

\begin{figure}[t]
\centering
\includegraphics[width=\linewidth]{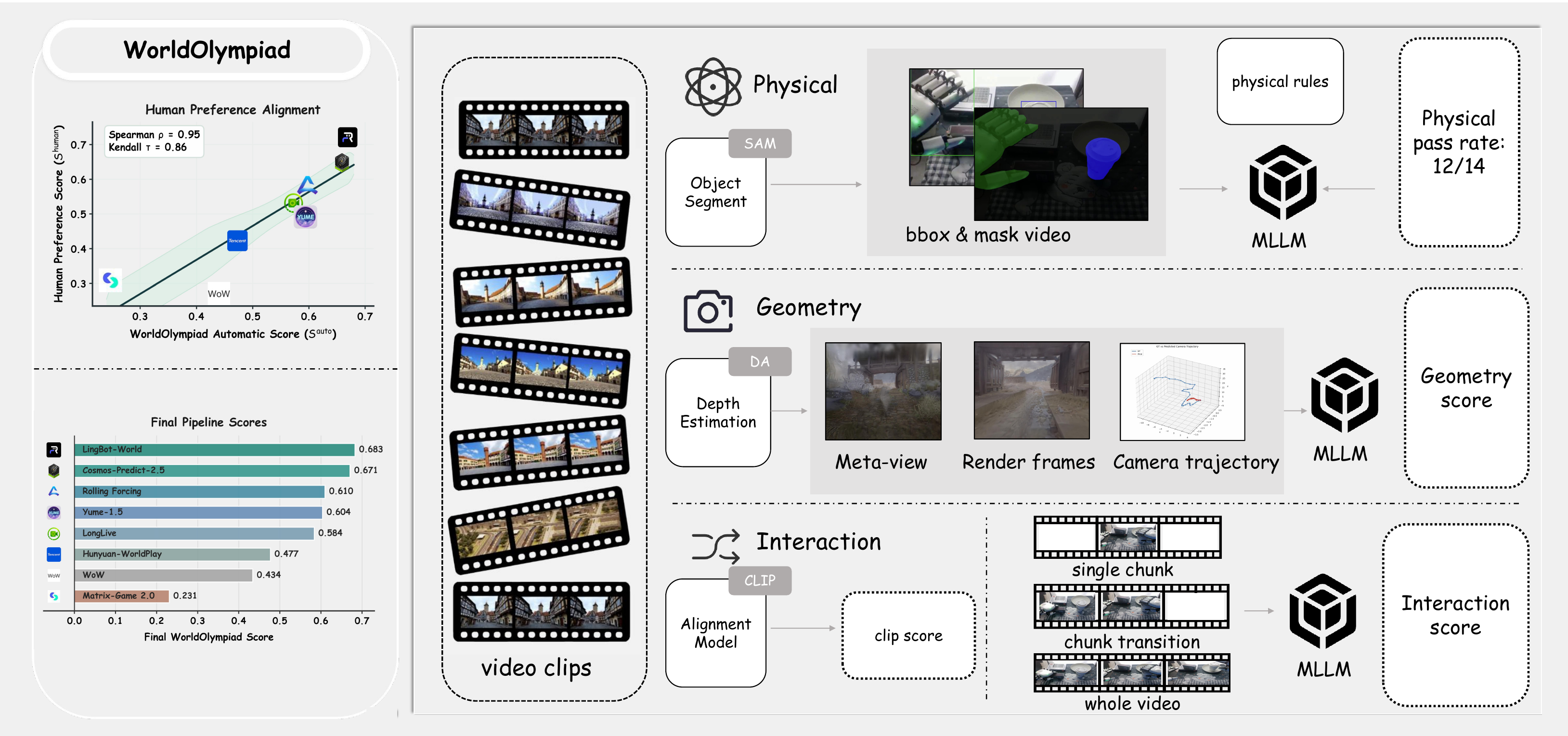}
\caption{Overview of the WorldOlympiad pipeline for data collection, long-video generation, and multi-dimensional evaluation.}
\label{fig:overview}
\end{figure}

Our contributions are summarized as follows:
\begin{itemize}
\item We propose \textbf{WorldOlympiad}, a unified benchmark for evaluating interactive long-video world models across gaming, robotics, and real-world scenarios.
\item We design multi-dimensional judge metrics that systematically assess physical-law adherence, 3D geometric consistency, and chunk-by-chunk interactive generation.
\item We construct a dataset of 1,000 high-quality long videos and benchmark 8 long-video generation pipelines, providing a systematic evaluation of their reliability in downstream world-model applications.
\end{itemize}

\section{Related Work}

\subsection{Video Generation}

Diffusion-based video generation models~\citep{blattmann2023stable,yang2024cogvideox,kong2024hunyuanvideo,wan2025wan,
team2025longcat} have demonstrated emergent physical consistency through
large-scale training~\citep{brooks2024video,wiedemer2025video}, including object permanence, 3D coherence, and plausible
motion dynamics. Despite these compelling properties, many early diffusion-based
video generators are optimized for short clips, often on the order of 5--10
seconds, which limits their direct use as persistent world model simulators.
Recently, block diffusion has emerged as a promising paradigm for scalable
long-horizon video synthesis. By performing iterative diffusion denoising
within each block and conditioning on previously generated content via
cross-block KV caching, this approach combines the high-quality parallel
generation of diffusion models with the sequential consistency of
autoregressive conditioning~\citep{huang2025self,yin2025slow,zhu2026causal,zhang2025blockvid}.
Such a design preserves intra-block denoising quality while enabling scalable
temporal extension, positioning block diffusion as a viable road toward
video-based world models.

\begin{table}[H]
\centering
\small
\caption{Comparison of existing benchmarks across evaluation metrics and video tasks.}
\label{tab:benchmark_comparison}
\setlength{\tabcolsep}{6pt}
\renewcommand{\arraystretch}{1.15}
\begin{tabular}{lccccccc}
\toprule
\multirow{2}{*}{\textbf{Benchmark}} 
  & \multicolumn{4}{c}{\textbf{Eval Metrics}} 
  & \multicolumn{3}{c}{\textbf{Video Tasks}} \\
\cmidrule(lr){2-5}\cmidrule(lr){6-8}
  & \makecell{Long Video} 
  & \makecell{Physical} 
  & \makecell{Geometry} 
  & \makecell{Interaction} 
  & \makecell{Gaming} 
  & \makecell{Robotics} 
  & \makecell{Real-world} \\
\midrule
VBench~\citep{huang2024vbench}        & \xmark & \xmark & \xmark & \xmark & \xmark & \xmark & \cmark \\
VBench++~\citep{huang2025vbench++}      & \cmark & \xmark & \xmark & \xmark & \xmark & \xmark & \cmark \\
VBench 2.0~\citep{zheng2025vbench}    & \xmark & \cmark & \xmark & \xmark & \xmark & \xmark & \cmark \\
MIND~\citep{ye2026mind}          & \cmark & \xmark & \xmark & \cmark & \cmark & \xmark & \xmark \\
EWMBench~\citep{hu2025ewmbench}        & \xmark & \xmark & \xmark & \cmark & \xmark & \cmark & \xmark \\
WorldEval~\citep{li2025worldevalworldmodelrealworld}       & \xmark & \xmark & \xmark & \cmark & \xmark & \cmark & \xmark \\
WorldArena~\citep{shang2026worldarena}    & \xmark & \cmark & \cmark & \cmark & \xmark & \cmark & \xmark \\
\midrule
\textbf{WorldOlympiad} & \cmark & \cmark & \cmark & \cmark & \cmark & \cmark & \cmark \\
\bottomrule
\end{tabular}
\end{table}

\subsection{Video Generation Models as World Models}

The rapid advancement of world models has enabled video generation to be deployed across diverse domains, including interactive game generation~\citep{team2026advancing} and robotics simulation~\citep{agarwal2025cosmos,ali2025world}. In the gaming domain, models such as GameGen-X~\citep{che2025gamegen} and Matrix Game~\citep{zhang2025matrix} have demonstrated compelling interactive game simulation with controllable character actions and environment dynamics. In robotics and embodied intelligence, dedicated interactive world models provide policy generation and data augmentation capabilities for robotic agents~\citep{agarwal2025cosmos,ali2025world,chi2025wow}. However, simultaneously maintaining persistent world state and supporting real-time interaction remains a significant challenge, giving rise to two core research directions. For memory and long-context modeling, some approaches adopt implicit memory mechanisms. For instance, LongLive~\citep{yang2025longlive} introduces KV caching to enable long-range consistent generation. In contrast, other works explicitly incorporate 3D memory mechanisms to preserve world-state consistency over extended horizons~\citep{xiao2025worldmem,huang2025memory,li2025vmem,zhao2025spatia,yu2026mosaicmem,wu2025video,wang2026mirage}. More recently, MosaicMem~\citep{yu2026mosaicmem} and Inspatio World~\citep{team2026inspatio} have begun exploring hybrid memory mechanisms, demonstrating substantial promise. On the interactive generation front, the dominant paradigm adapts controllable video generation techniques within the block diffusion framework~\citep{liu2026realwonder,shin2025motionstream}, enabling interactive video synthesis. Works such as LingBot-World~\citep{team2026advancing} have shown strong performance on downstream tasks such as interactive game generation through scaling. Regardless of the target application, real-time interaction remains a central capability requirement in this field.

\subsection{World Model Benchmarks}

Existing benchmarks for short video generation have introduced a broad set of
general evaluation metrics, as exemplified by VBench~\citep{huang2024vbench} and its
successor VBench 2.0~\citep{zheng2025vbench}, which cover multi-dimensional criteria
spanning visual quality, motion authenticity, semantic consistency, and
physical plausibility. More recently, benchmarks specifically targeting world
model capabilities have been proposed, evaluating models along dimensions such
as physical law adherence, simulation fidelity, and functional world modeling~\citep{duan2025worldscore,qin2024worldsimbench,li2025worldmodelbench, ying2026wbenchcomprehensivemultiturnbenchmark}. Newer benchmarks tailored to
robotics downstream tasks~\citep{shang2026worldarena,deng2026rethinking,liu2026rise} extend the evaluation scope to
controllability, action conditioning, and closed-loop interaction. Despite this
progress, existing benchmarks lack unified coverage across multiple downstream
application domains, including gaming, robotics, and general scene generation,
within a single evaluation framework. Moreover, assessments of interactive
functionality, which is arguably the most critical capability of world models,
remain notably absent. To address these gaps, we propose WorldOlympiad, a
comprehensive benchmark that unifies the evaluation of game, robotics, and
real-world environments, comprising 1,000 high-quality video samples
spanning diverse downstream scenarios, and jointly assessing perceptual quality
alongside functional world modeling capabilities.

\section{WorldOlympiad}
\subsection{Data Collection}

\begin{figure}[tbp]
\centering
\includegraphics[width=\linewidth]{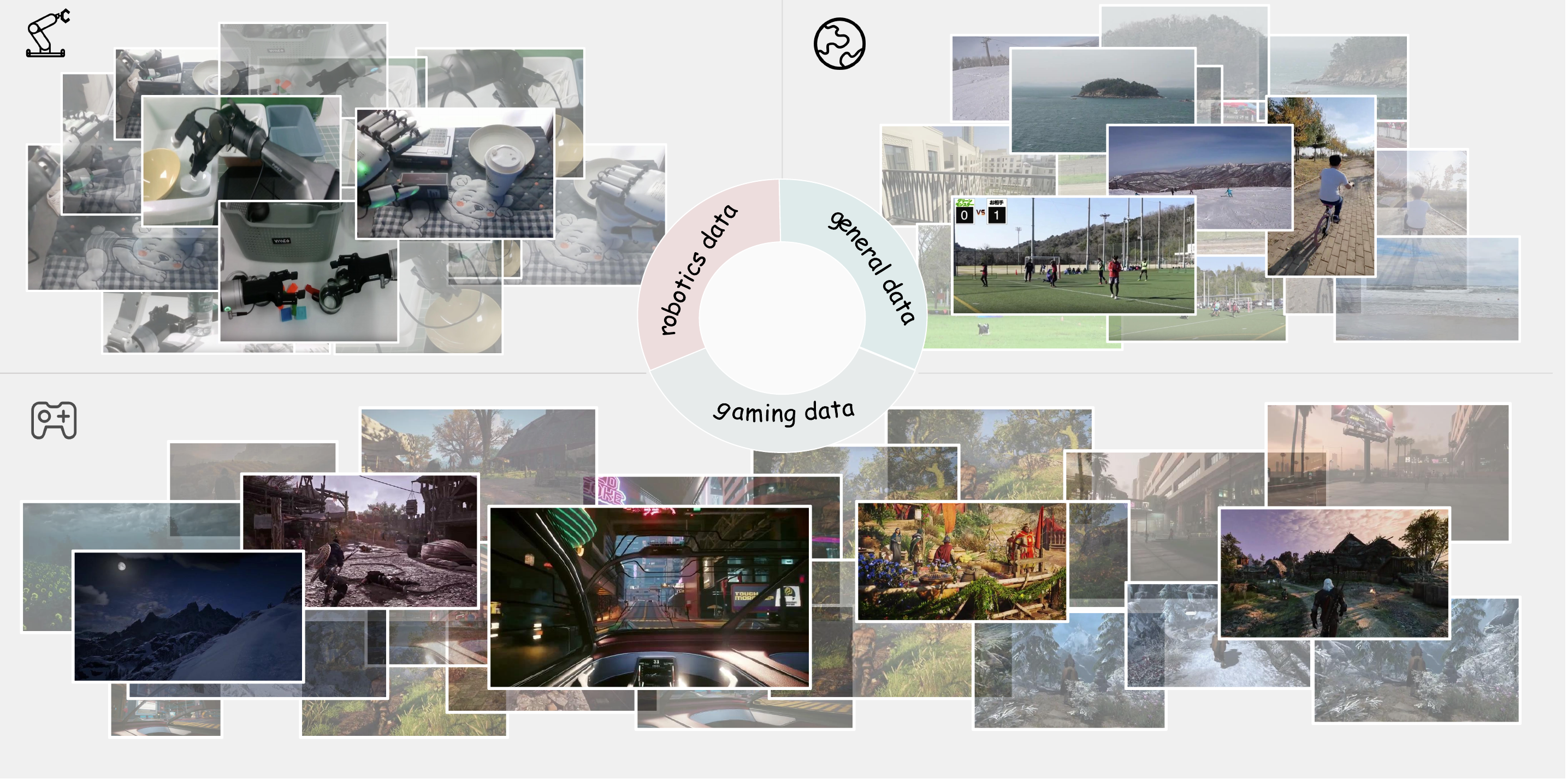}
\caption{Data collection overview across robotics, gaming, and real-world video sources.}
\label{fig:data_overview}
\end{figure}

Figure~\ref{fig:data_overview} summarizes the data collection process of WorldOlympiad. The benchmark contains 400 robotics videos, 400 gaming videos, and 200 real-world videos, covering complementary world-modeling requirements: robotics videos emphasize object manipulation and physical interaction, gaming videos emphasize interactive control and long-context state evolution, and real-world videos emphasize open-domain motion and camera dynamics. This diverse composition enables a comprehensive evaluation of video-based world models across their three most critical application domains.

\subsubsection{Source Domains}

\paragraph{Robotics Domain.} The robotics subset is built from RoboCOIN~\citep{wu2025robocoin}, an open-source bimanual robotic manipulation data collection. We use this source because bimanual manipulation naturally contains object contact, gripper motion, state changes, and physically grounded interactions. RoboCOIN also includes multiple bimanual robot embodiments, giving the subset broad coverage for evaluating whether generated videos preserve action-consistent dynamics. From the downloaded RoboCOIN videos, we manually filter 400 videos as the robotics portion of the benchmark test set.

\paragraph{Gaming Domain.} The gaming subset is built from GameGen-X~\citep{che2025gamegen}, an interactive open-world game video dataset. We randomly sample videos from the official \texttt{OGameData\_50K.csv} metadata file and download the corresponding videos. Since some gameplay videos are usually long and contain multiple interaction stages, we split long videos into shorter video chunks with 60 seconds before constructing the final 400-video gaming subset. This subset targets interactive world-modeling behavior such as camera movement, player navigation, combat events, skill execution, and game-state changes.

\paragraph{Real-world Domain.} The real-world subset is built from LVD-2M~\citep{xiong2024lvd}, a long-take video dataset with temporally dense captions. We use the official \texttt{ytb\_600k\_720p.csv} subset and randomly select videos whose duration is longer than 60 seconds and whose motion score is greater than 50. This filtering rule favors long videos with sufficient visible motion, making the subset suitable for evaluating open-domain dynamics, camera movement, and geometric consistency in everyday scenes.

\begin{table}[tbp]
\centering
\small
\caption{Data composition of the WorldOlympiad benchmark test set.}
\label{tab:data_composition}
\setlength{\tabcolsep}{5pt}
\renewcommand{\arraystretch}{1.15}
\begin{tabular}{lclp{6.2cm}}
\toprule
\textbf{Domain} & \textbf{Count} & \textbf{Source} & \textbf{Selection rule} \\
\midrule
Robotics & 400 & RoboCOIN~\citep{wu2025robocoin} & Downloaded videos that are manually filtered. \\
Gaming & 400 & GameGen-X~\citep{che2025gamegen} & Randomly sampled videos from the official \texttt{OGameData\_50K.csv}; long videos are split into shorter evaluation chunks. \\
Real-world & 200 & LVD-2M~\citep{xiong2024lvd} & Videos selected from \texttt{ytb\_600k\_720p.csv} with duration longer than 60 seconds and motion score greater than 50. \\
\bottomrule
\end{tabular}
\end{table}

\FloatBarrier

\subsubsection{Temporal Chunking and Captioning}
\begin{figure}[t]
\centering
\includegraphics[width=\linewidth]{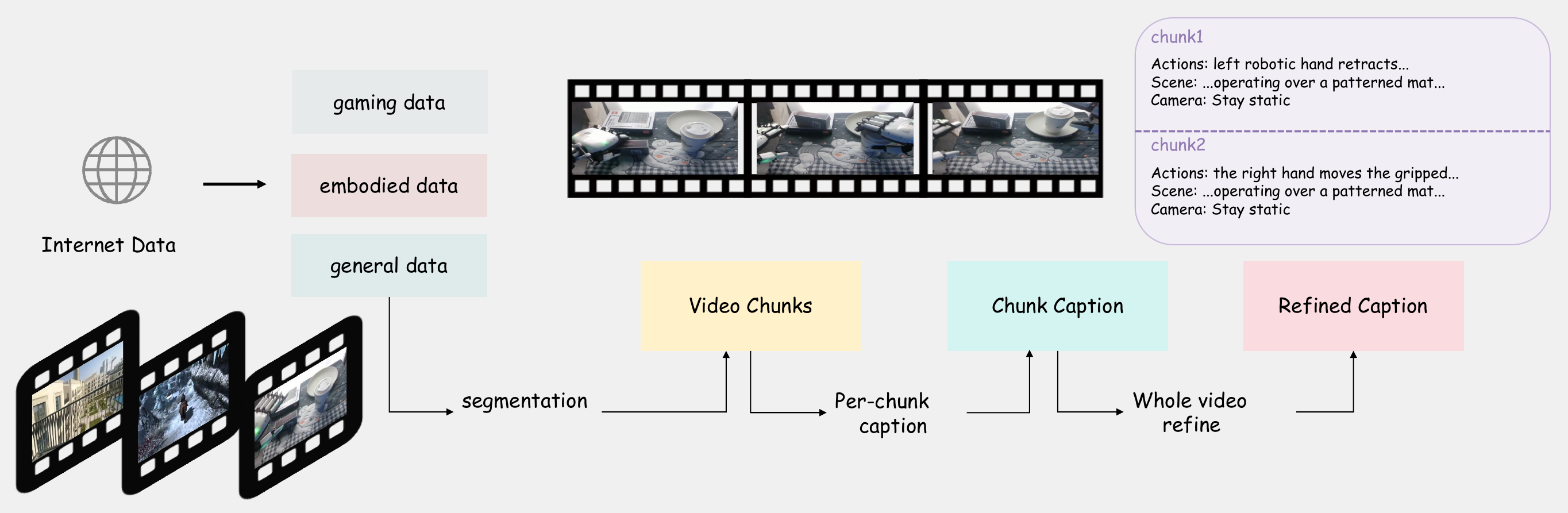}
\caption{Data standardization pipeline from raw videos to refined action-caption annotations.}
\label{fig:data_composition_pipeline}
\end{figure}

Detailed video captions are essential for subsequent evaluation. Instead of relying on a single-pass MLLM, we design a three-stage chunk-caption-refine pipeline to ensure the resulting annotations are both accurate and comprehensive, as illustrated in Figure~\ref{fig:data_composition_pipeline}. We adopt Gemini-3-Pro-Preview~\citep{gemini3} across all stages, owing to its superior performance in multimodal understanding.


\paragraph{\textbf{StageI-Chunking.}} The pipeline first identifies the main continuous execution interval in a video and divides it into at most six contiguous chunks. All chunks follow a left-closed, right-open interval convention, and adjacent chunks are required to have no temporal gaps or overlaps. For gaming videos, the chunking prompt focuses on gameplay execution such as combat, traversal, skill casting, and camera transitions; for real-world videos, the prompt focuses on continuous visual actions, object motion, interaction events, and view transitions. 

\paragraph{\textbf{StageII-Caption.}} After temporal chunking, the captioning model generates chunk-level captions for each video chunk. For each robotics, gaming, or real-world chunk, the captioning model outputs two fields: an \texttt{action} field and a \texttt{caption} field. The \texttt{action} field maps camera movement to WASD-style controls, with \texttt{None} used when the camera does not move noticeably. This action label is intentionally based on camera movement only; it is not inferred from character animation, subject motion, visual effects, or UI changes. The \texttt{caption} field describes the scene, visible entities, events, interactions, and outcomes in English.

\paragraph{\textbf{StageIII-Refine.}} We then refine the chunk-level captions with the full video as context. Given the full video and the time-ordered chunk captions, the refinement step corrects hallucinated details, standardizes terminology across adjacent chunks, improves narrative continuity, and validates the camera-movement action label. This final pass is important for long-video evaluation because adjacent chunks often share objects, locations, player states, or scene context, and inconsistent captions would weaken the reliability of interaction and long-context assessment.

We take the outputs from Stage III as the final captions for each video chunk, which are subsequently used for evaluation. The active judge prompts used by WorldOlympiad are provided in Appendix~\ref{app:prompt-templates}.


\subsection{Evaluation Metrics}


\subsubsection{Physical Evaluation}

We evaluate physical faithfulness with a rule-based benchmark spanning three
subsets: mechanics, thermodynamics, and material properties. The pipeline first uses an
MLLM to identify the moving or deforming entities that are most
relevant to physical reasoning, and then applies SAM3~\citep{carion2025sam} to produce
object-centric visualizations that expose their masks and trajectories more
clearly. After this preprocessing stage, each metric is evaluated in two steps.
A relevance judge first determines whether the target phenomenon is actually
present in the ground-truth reference video under the given prompt; unrelated metrics are
marked as not related and excluded from scoring. For each relevant metric, a
compliance judge then compares the generated video with a ground-truth
reference video and predicts whether the observed behavior follows the
corresponding physical rule, together with a confidence score and a short
explanation. Final physical results are reported by averaging compliance over
the applicable metrics within each subset and then across subsets. The active
mechanics, thermodynamics, and material rule prompt templates used by the
physical MLLLM judges are provided in Appendix~\ref{app:physical-judge-prompts}.

\paragraph{\textbf{Mechanics.}}
\textit{Gravity} evaluates whether unsupported objects move downward under gravity, rather than floating upward or accelerating in physically implausible directions. \textit{Buoyancy} focuses on whether objects in fluids remain near the surface or sink in accordance with their apparent density. \textit{Compression} measures whether solids deform plausibly under load, instead of staying unrealistically rigid or buckling without sufficient cause. \textit{Impact} examines whether collisions lead to reasonable post-impact dynamics, including momentum transfer, rebound, fracture, or eventual rest.

\begin{figure}[H]
\centering
\includegraphics[width=\linewidth]{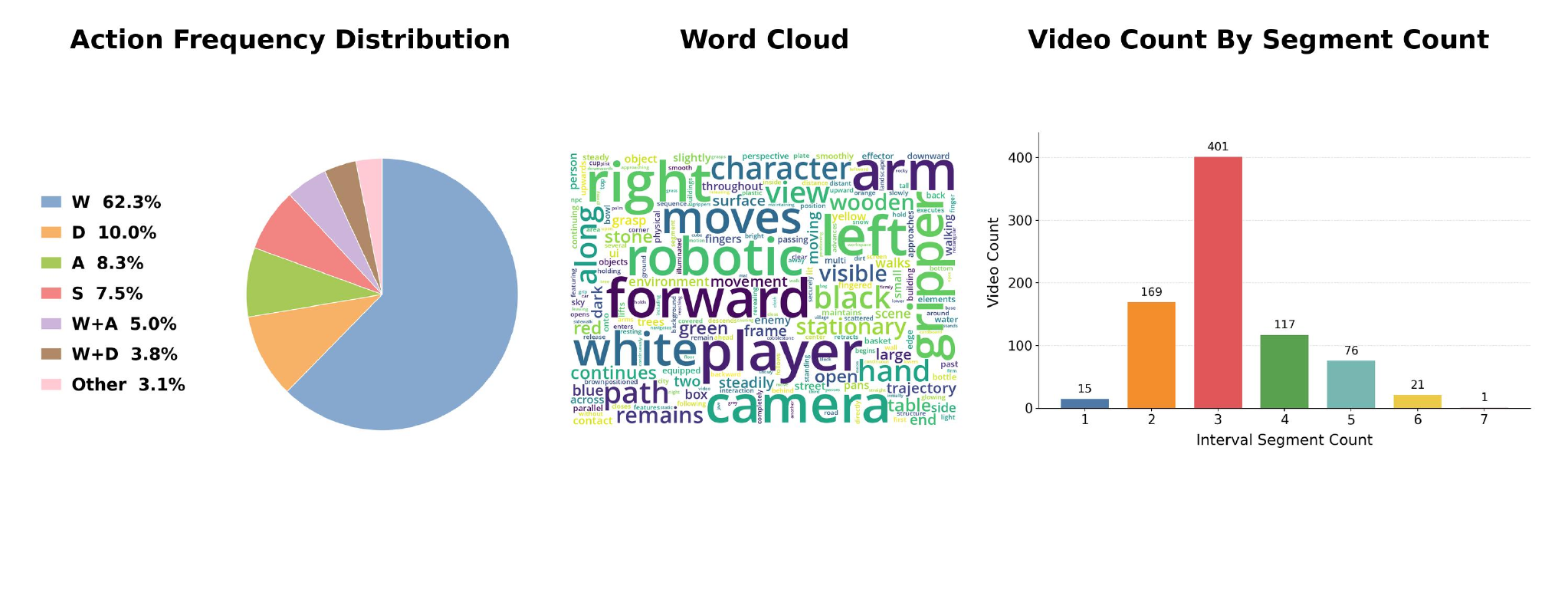}
\caption{Pipeline statistics for data processing, annotation coverage, and evaluation-ready samples.}
\label{fig:pipeline_statistics}
\end{figure}

\paragraph{\textbf{Thermodynamics.}}
\textit{Melting} assesses whether a heated solid gradually transitions into a liquid state. \textit{Sublimation} captures direct solid-to-gas transitions without an intermediate liquid phase. \textit{Vaporization} considers whether liquids turn into vapor through evaporation or boiling when heated or exposed over time. \textit{Condensation} evaluates the formation of liquid droplets from cooled gas. \textit{Deposition} describes the direct transformation from gas to solid without first becoming liquid. \textit{Freezing} measures whether a cooled liquid solidifies into a stable solid state.

\paragraph{\textbf{Material.}}
\textit{Color mixing} evaluates whether mixed colored liquids or paints yield the expected resultant color. \textit{Solubility} focuses on whether soluble substances disperse and dissolve into the solvent, rather than remaining intact. \textit{Hardness} distinguishes whether soft materials bend or tear easily while hard materials resist deformation or break sharply. \textit{Combustibility} examines whether flammable materials ignite and produce physically consistent fire, smoke, or charring behavior.

\subsubsection{Geometry Evaluation}

We evaluate geometric consistency with three complementary signals~\citep{wang2026world}:
\textit{\(S_{\mathrm{recon}}\)} scores the rendered Gaussian-Splat video,
\textit{\(S_{\mathrm{meta}}\)} scores a diagnostic meta-view, and
\textit{\(S_{\mathrm{traj}}\)} scores agreement between the recovered and
reference camera trajectories. Given a generated video
\(V=\{I_t\}_{t=1}^{T}\), we uniformly sample
\(\bar V=\{I_i\}_{i=1}^{N}\), with \(N\leq 32\) in the implementation. When
dynamic-object masks are available, foreground Gaussians are removed before
rendering so that the 3D judge focuses on the static scene. Depth Anything
3~\citep{lin2025depth} estimates a Gaussian scene and camera parameters, and
the Gaussian-Splat renderer produces two diagnostic artifacts:
\begin{equation}
\mathcal{F}_{\mathrm{DA3}}(\bar V)\rightarrow
\left(\mathcal{G},\{E_i,K_i\}_{i=1}^{N}\right),\quad
\hat V_{\mathrm{GS}}=\mathcal{R}(\mathcal{G},\{E_i,K_i\}_{i=1}^{N}),\quad
\hat I_{\mathrm{meta}}=\mathcal{R}(\mathcal{G},E_{i^\star},K_{i^\star}),
\end{equation}
where \(\mathcal{G}\) is the reconstructed Gaussian representation,
\(E_i\) and \(K_i\) are recovered extrinsics and intrinsics, and \(i^\star\)
denotes the recovered camera pose farthest from the reconstruction origin.

The reconstruction and meta-view scores are produced by the same calibrated
MLLM judge used in the implementation. The judge inspects whether the rendered
static scene preserves a recognizable layout, coherent 3D structure, stable
cross-view geometry, and prompt-consistent scene organization. The judge is
instructed to return a strict JSON score in \([0,1]\), and the parsed scores
are clamped to \([0,1]\) to avoid ambiguity with the CLIP model used in the
interaction metric:
\begin{equation}
S_{\mathrm{recon}}=\operatorname{clamp}\!\left(J_{\mathrm{vid}}(\hat V_{\mathrm{GS}},p),0,1\right),
\qquad
S_{\mathrm{meta}}=\operatorname{clamp}\!\left(J_{\mathrm{img}}(\hat I_{\mathrm{meta}},p),0,1\right),
\end{equation}
where \(p\) is the static-scene prompt used for 3D judging. In the optional
LPIPS setting, the Gaussian-Splat video score is replaced by
\(\operatorname{clamp}(1-\mathrm{LPIPS}(\hat V_{\mathrm{GS}},\bar V),0,1)\).

For camera motion, let \(\{\hat T_i\}_{i=1}^{L}\) and \(\{T_i\}_{i=1}^{L}\)
denote the predicted and reference camera-to-world trajectories after temporal
resampling to a shared length. If the reference contains non-negligible
translation, the predicted trajectory is first aligned to the reference by a
similarity transform. Both trajectories are then expressed relative to their
first frame:
\begin{equation}
\tilde T_i=T_1^{-1}T_i,\qquad
\tilde{\hat T}_i=\hat T_1^{-1}\hat T_i,\qquad i=1,\ldots,L .
\end{equation}
The translation score \(S_t\) combines path-shape similarity, motion-extent
agreement, and mean camera-center error. The rotation score \(S_r\) combines
mean geodesic rotation error, final-frame rotation error, and total
rotation-extent agreement. The final trajectory score is computed by an
adaptive aggregation function \(A_{\mathrm{motion}}\):
\begin{equation}
S_{\mathrm{traj}} =
A_{\mathrm{motion}}\left(S_t,S_r;\{\tilde T_i\}_{i=1}^{L}\right).
\end{equation}
This aggregation is selected from the reference motion profile. For nearly
static trajectories, the score penalizes reconstructed camera jitter directly.
For translation-dominant or rotation-dominant trajectories, the corresponding
component receives the larger weight; for mixed motion, translation and
rotation are weighted evenly.

The implementation records the raw 3D reward as the sum of the three bounded
subscores, while all tables report the normalized geometry score:
\begin{equation}
S_{3D}=\frac{1}{3}\left(S_{\mathrm{recon}}+S_{\mathrm{meta}}+S_{\mathrm{traj}}\right).
\end{equation}

\subsubsection{Interaction Evaluation}

We evaluate interaction fidelity under the chunk-by-chunk generation setting.
Given a generated video divided into \(T\) chunks
\(\{v_i\}_{i=1}^{T}\) and their corresponding captions
\(\{p_i\}_{i=1}^{T}\), the interaction benchmark measures whether each chunk
follows its local instruction, whether adjacent chunks transition coherently,
and whether the full video remains temporally fluent. This design matches the
way interactive video world models are typically rolled out: each new chunk is
conditioned on the previous visual context and a new control or action caption,
so a model must satisfy both local caption alignment and long-range continuity.

The first component is a CLIP-based semantic-adherence score. For each chunk,
we uniformly sample a fixed number of frames within its temporal interval
\(F_i=\{f_{i,j}\}_{j=1}^{m_i}\), where \(m_i\) is 8 by default. We encode each
sampled frame and the corresponding chunk caption with a CLIP model~\citep{radford2021learning,yang2025longlive}, convert both embeddings to unit-length vectors, and compute their dot-product similarity. The
chunk-level score is the mean similarity over sampled frames,
\begin{equation}
s_i^{\mathrm{clip}} = \frac{1}{m_i}\sum_{j=1}^{m_i}
\mathrm{sim}\bigl(\mathrm{CLIP}_{v}(f_{i,j}), \mathrm{CLIP}_{t}(p_i)\bigr),
\end{equation}
and the video-level semantic-adherence score is the weighted mean over all
valid sampled frames:
\begin{equation}
S_{\mathrm{clip}} =
\frac{\sum_{i=1}^{T}\sum_{j=1}^{m_i}
\mathrm{sim}\bigl(\mathrm{CLIP}_{v}(f_{i,j}), \mathrm{CLIP}_{t}(p_i)\bigr)}
{\sum_{i=1}^{T}m_i}.
\end{equation}
Because this is a cosine similarity computed from normalized CLIP embeddings,
the raw score remains on its native \([-1,1]\) scale. To use it as a bounded
auxiliary interaction signal, we convert it into a calibrated semantic score
with fixed thresholds:
\begin{equation}
\widetilde{S}_{\mathrm{clip}} =
\operatorname{clip}\left(
\frac{S_{\mathrm{clip}}-\tau_{\min}}{\tau_{\max}-\tau_{\min}},
0, 1
\right),
\quad \tau_{\min}=0.20,\quad \tau_{\max}=0.40 .
\end{equation}
The thresholds are fixed across all evaluated models, so adding a new model
does not change previously reported CLIP auxiliary scores. This component
provides an automatic and lightweight estimate of whether the generated chunks
contain the semantic content requested by their captions.

The second component uses an MLLM as a structured rubric-based judge. We query
the MLLM at three complementary levels, and all returned scores are clipped to
the requested 0--5 range before being normalized to \([0,1]\) for reporting. First, the MLLM
receives each chunk \(v_i\) and its caption \(p_i\), and scores
\textit{visual quality}, \textit{text alignment}, and an \textit{overall}
chunk score \(a_i\). Second, the MLLM receives each adjacent pair
\((v_i, v_{i+1})\) together with their captions \((p_i, p_{i+1})\), and scores
\textit{transition smoothness} and an \textit{overall} transition score
\(b_i\). Third, the MLLM receives the full generated video and scores
\textit{long-range consistency}, \textit{global text alignment}, and a global
\textit{overall} score \(g\). The final MLLM interaction score averages the
overall scores from the chunk, transition, and global judgments:
\begin{equation}
S_{\mathrm{chunk}} = \frac{1}{5T}\sum_{i=1}^{T} a_i,\quad
S_{\mathrm{trans}} = \frac{1}{5(T-1)}\sum_{i=1}^{T-1} b_i,\quad
S_{\mathrm{global}} = \frac{g}{5},
\end{equation}
\begin{equation}
S_{\mathrm{mllm}} =
\frac{1}{3}\left(S_{\mathrm{chunk}} + S_{\mathrm{trans}} +
S_{\mathrm{global}}\right).
\end{equation}

The final interaction score uses the calibrated CLIP score as a lightweight
semantic auxiliary term:
\begin{equation}
S_{\mathrm{interact}} =
(1-\lambda)S_{\mathrm{mllm}} + \lambda\widetilde{S}_{\mathrm{clip}},
\quad \lambda=0.1 .
\end{equation}
This design lets CLIP contribute frame-caption semantic grounding while keeping
the interaction metric dominated by the structured MLLM judge, which evaluates
temporal properties such as chunk-level instruction following, boundary
smoothness, state preservation, and full-video fluency.

Finally, WorldOlympiad reports an overall score by averaging the three core
evaluation tracks:
\begin{equation}
S_{\mathrm{all}} =
\frac{1}{3}\left(S_{\mathrm{phys}} + S_{3D} + S_{\mathrm{interact}}\right).
\end{equation}
This equal-weight aggregation keeps the leaderboard aligned with the benchmark
design: physical faithfulness, geometric consistency, and interaction fidelity
contribute symmetrically to the final model ranking.

\section{Experiment}

\subsection{Experimental Setup}

\paragraph{\textbf{Evaluation models.}}
We evaluate eight publicly available video-generation pipelines through
OpenWorldLib~\citep{team2026openworldlib}. These pipelines cover three major
families of video world models. The gaming-centric group includes
Matrix-Game 2.0~\citep{he2025matrix} and
LingBot-World~\citep{team2026advancing}; the robotics-centric group includes
Cosmos-Predict-2.5~\citep{ali2025world} and WoW~\citep{chi2025wow}; and the
general long-video group includes Rolling Forcing~\citep{liu2025rolling},
LongLive~\citep{yang2025longlive}, Yume-1.5~\citep{mao2025yume15}, and
Hunyuan-WorldPlay~\citep{sun2025worldplay}. In our experiments, we test these
pipelines across different downstream scenarios, including gaming, robotics,
and general real-world videos.

\paragraph{\textbf{Implementation details.}}
For fairness, we use each released pipeline with its official default
generation configuration whenever possible. Since different pipelines may adopt
different chunk sizes or segment-level generation settings, we dynamically map
the temporal information in the chunk captions to each model's native generation
configuration. This allows the temporal proportions of the original chunk captions to be retained while respecting each generation pipeline's native training and inference configuration.
 For methods that include an
explicit memory or long-context mechanism, such as Rolling Forcing, we preserve
the official memory-management strategy during rollout. For pipelines without a
dedicated long-horizon memory module, such as WoW, we perform long-video
generation through video continuation, using the previously generated context
as the condition for the next segment.

All generated videos are evaluated by the same automatic WorldOlympiad
evaluator. The evaluator reports physical faithfulness, 3D consistency,
CLIP-augmented interaction fidelity, and an overall composite score.
Judge-based component scores are reported after averaging their normalized
subscores into the $[0,1]$ range. Physical faithfulness aggregates rule-level
judgments over mechanics, thermodynamics, and material behavior; 3D consistency
combines reconstruction quality, meta-view quality, and camera trajectory
consistency; and interaction fidelity measures chunk-level instruction
following, CLIP-based semantic grounding, adjacent transition smoothness, and
long-range coherence over the full generated video.

\subsection{Main Benchmark Results}

\begin{figure}[tbp]
\centering
\includegraphics[width=\linewidth]{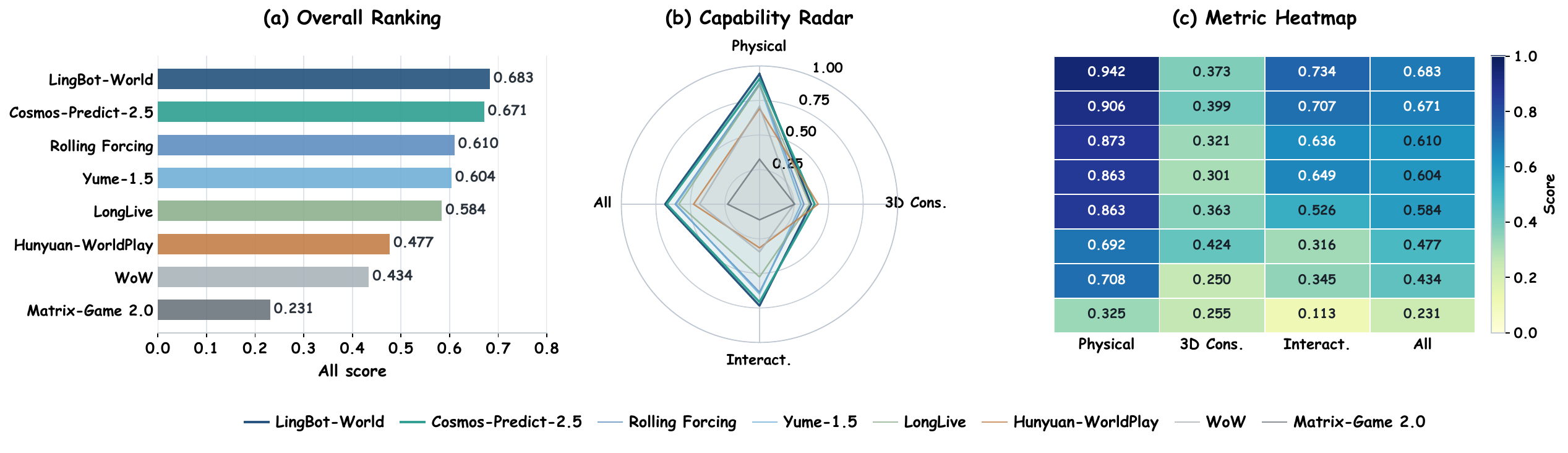}
\caption{Result statistics of WorldOlympiad across evaluated world-model pipelines and scoring dimensions.}
\label{fig:result_statistics}
\end{figure}

Table~\ref{tab:openworldlib_eval} summarizes the video world models evaluated
in OpenWorldLib, grouped by gaming, robotics, and general world-model
categories. The table reports physical faithfulness, 3D consistency,
CLIP-augmented interaction fidelity, and the overall score.
Figure~\ref{fig:result_statistics} further visualizes the score distribution
across pipelines and evaluation dimensions.

\begin{table*}[t]
\centering
\small
\caption{
    Main benchmark results on \textbf{WorldOlympiad}. 
    We evaluate eight representative video world models across gaming, robotics, and general long-video generation settings.
    \textbf{Physical} (\(S_{\mathrm{phys}}\)): physical faithfulness;
    \textbf{3D Cons.} (\(S_{3D}\)): 3D spatial consistency;
    \textbf{Interact.} (\(S_{\mathrm{interact}}\)): interaction fidelity with CLIP-based semantic grounding;
    \textbf{All} (\(S_{\mathrm{all}}\)): overall composite score.
    Best and second-best results are marked in \textbf{bold} and \underline{underlined}, respectively.
}
\label{tab:openworldlib_eval}
\setlength{\tabcolsep}{7pt}
\renewcommand{\arraystretch}{1.22}
\resizebox{\textwidth}{!}{
\begin{tabular}{llccccc}
\toprule
\multirow{2}{*}{\textbf{Category}} 
& \multirow{2}{*}{\textbf{Model}}
& \multicolumn{4}{c}{\textbf{Evaluation Metrics}}
& \multirow{2}{*}{\textbf{Rank}} \\
\cmidrule(lr){3-6}
& 
& \textbf{Physical} 
& \textbf{3D Cons.} 
& \textbf{Interact.} 
& \textbf{All}
& \\
\midrule

\multirow{2}{*}{\makecell[l]{\textsc{Gaming}\\\textsc{World Model}}}
& Matrix-Game 2.0~\citep{he2025matrix}     
& 0.325 
& 0.255 
& 0.113 
& 0.231 
& 8 \\
& LingBot-World~\citep{team2026advancing}  
& \textbf{0.942} 
& 0.373 
& \textbf{0.734} 
& \textbf{0.683} 
& \textbf{1} \\

\midrule

\multirow{2}{*}{\makecell[l]{\textsc{Robotics}\\\textsc{World Model}}}
& Cosmos-Predict-2.5~\citep{ali2025world}  
& \underline{0.906} 
& \underline{0.399} 
& \underline{0.707} 
& \underline{0.671} 
& \underline{2} \\
& WoW~\citep{chi2025wow}                   
& 0.708 
& 0.250 
& 0.345 
& 0.434 
& 7 \\

\midrule

\multirow{4}{*}{\makecell[l]{\textsc{General}\\\textsc{World Model}}}
& Rolling Forcing~\citep{liu2025rolling}   
& 0.873 
& 0.321 
& 0.636 
& 0.610 
& 3 \\
& LongLive~\citep{yang2025longlive}         
& 0.863 
& 0.363 
& 0.526 
& 0.584 
& 5 \\
& Yume-1.5~\citep{mao2025yume15}           
& 0.863 
& 0.301 
& 0.649 
& 0.604 
& 4 \\
& Hunyuan-WorldPlay~\citep{sun2025worldplay} 
& 0.692 
& \textbf{0.424} 
& 0.316 
& 0.477 
& 6 \\

\bottomrule
\end{tabular}
}
\vspace{2pt}
\caption*{\footnotesize 
    \textbf{All} is the average of \textbf{Physical}, \textbf{3D Cons.}, and \textbf{Interact.}; overall ranks are computed by the unrounded \textbf{All} score. 
    Displayed scores are rounded to three decimal places.
}
\end{table*}

\paragraph{\textbf{From visual synthesis to stateful world simulation.}}
The most salient trend in Table~\ref{tab:openworldlib_eval} is that the best
models are no longer distinguished only by visual plausibility, but by their
ability to preserve physical state and interaction semantics over extended
rollouts. LingBot-World achieves the highest overall score (0.683), with
particularly strong physical faithfulness (0.942) and interaction fidelity
(0.734). Notably, LingBot-World is a 14B-activated model, suggesting that
large-scale capacity can substantially improve long-horizon state preservation,
scene continuity, and action-conditioned dynamics. However, model scale is not
the only factor that determines world-model quality. Cosmos-Predict-2.5, with
only 2B parameters, reaches a comparable overall score of (0.671).
Although it is categorized as a robotics-centric pipeline in our evaluation,
Cosmos-Predict-2.5 is optimized for physical-world prediction, which helps it
generalize beyond embodied manipulation scenarios and achieve strong physical
fidelity across diverse downstream settings. This comparison suggests that
targeted physical-world training and rollout design can partly compensate for
smaller activated model scale, leading to competitive performance in stateful
world simulation.

\paragraph{\textbf{Physical regularity is emerging as a shared capability.}}
A second trend is that several recent pipelines already show strong compliance
with common physical regularities. LingBot-World (\(0.942\)),
Cosmos-Predict-2.5 (\(0.906\)), Rolling Forcing (\(0.873\)),
LongLive (\(0.863\)), and Yume-1.5 (\(0.863\)) all achieve high physical scores,
suggesting that current video world models have begun to internalize frequent
patterns of motion, contact, support, and material behavior. This progress is
consistent with the increasing attention to physical plausibility in recent
evaluation suites such as VBench 2.0. However, the capability is still uneven:
fine-grained results in the appendix show that thermodynamics and material-level
questions remain more fragile than many mechanics questions, and weaker models
still violate basic constraints under long-horizon generation.

\paragraph{\textbf{The geometry-simulation gap remains unresolved.}}
Geometric consistency remains one of the most important unresolved weaknesses
across current video world models. Even the strongest pipeline on this
dimension, Hunyuan-WorldPlay, reaches only (0.424), while most models remain
in the (0.25)--(0.40) range. Notably, models represented by
Hunyuan-WorldPlay rely more heavily on camera or viewpoint control as their
primary form of interaction. This design encourages the model to preserve
spatial layout under view changes, which helps explain its relatively stronger
3D consistency. However, such interaction is also more constrained than
open-ended action-conditioned generation: controlling the camera or viewpoint
does not necessarily require the model to reason about complex object
manipulation, agent behavior, or multi-step state transitions. As a result,
these models can obtain better geometry scores while still achieving limited
overall performance. This highlights a key trade-off in current world models:
view-control pipelines may better preserve cross-view structure, but robust
world simulation requires both stable 3D geometry and flexible interactive
dynamics.

\paragraph{\textbf{The specialization-generalization trade-off.}}
LingBot-World and Cosmos-Predict-2.5 have both undergone sustained training in specific domains such as gaming and robotics. Their strong performance in our benchmark suggests that continuous domain-specific training can effectively generalize to broader evaluation settings. In particular, the fact that these two specialized pipelines rank at the top indicates that targeted training does not necessarily limit a model to its original domain; instead, it can provide transferable world knowledge that benefits performance across different scenarios.However, not all specialized models show the same generalization ability. WoW performs better in embodied scenarios than in other domains, but its scores drop on gaming and general real-world videos. As shown in Table~6, WoW reaches $0.502$ on embodied videos, but only $0.368$ on gaming videos and $0.415$ on general videos. These results suggest that specialization is useful only when the learned knowledge can transfer beyond a narrow domain. Future models should therefore combine sustained domain-specific training with broader cross-domain world knowledge.

\paragraph{\textbf{Fine-grained diagnostics.}}
WorldOlympiad is designed to be diagnostic rather than only leaderboard-driven.
Beyond the aggregate scores in Table~\ref{tab:openworldlib_eval}, we decompose
model behavior into domain-level results, physical dimensions and questions,
3D reconstruction submetrics, and interaction submetrics. These breakdowns
make it possible to identify whether a low score is caused by a specific
physical rule, unstable geometry, poor semantic grounding, or long-range
interaction drift. Detailed tables for these fine-grained results are provided
in Appendix~\ref{app:detailed-results}.

\begin{figure}[tbp]
\centering
\includegraphics[width=\linewidth]{{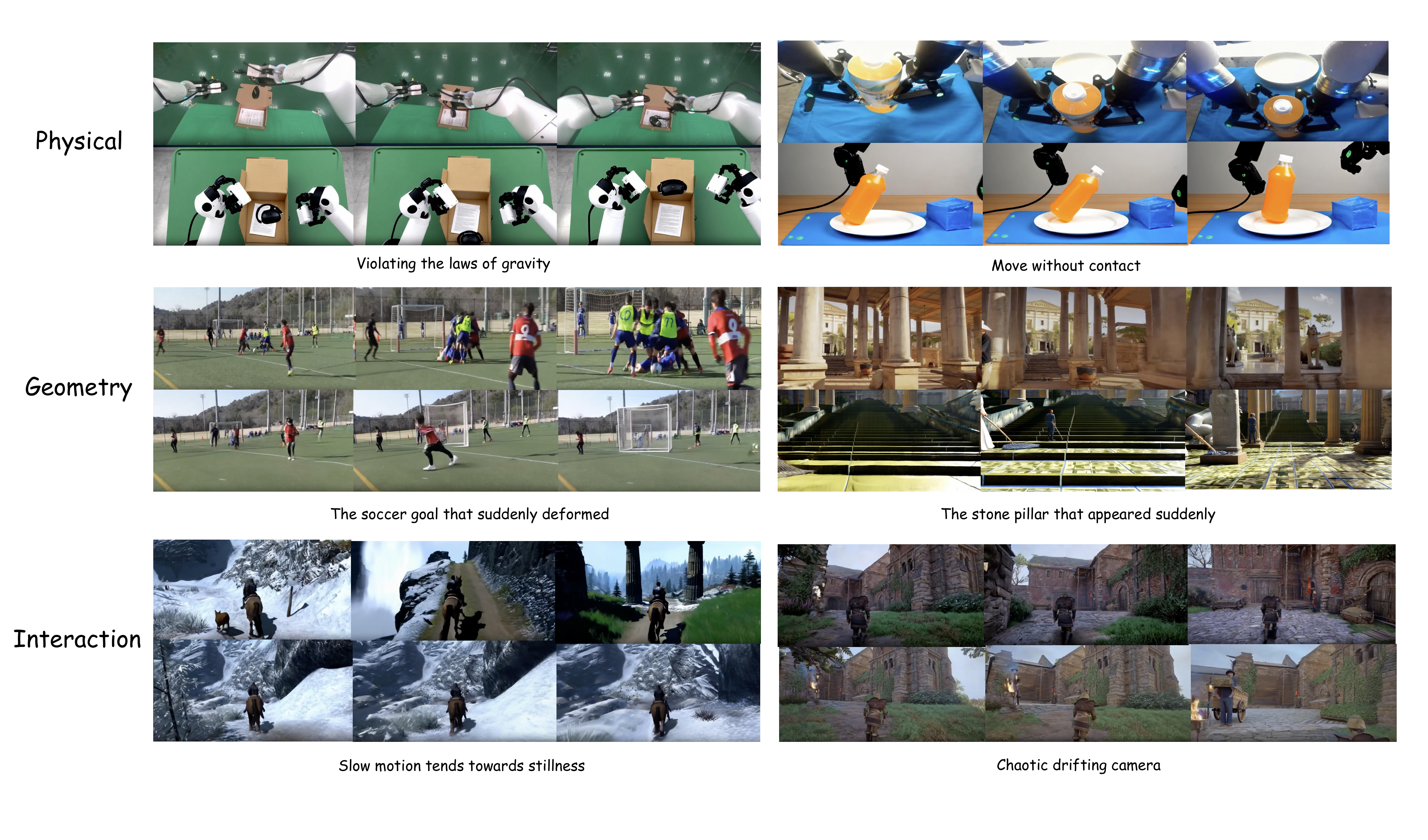}}
\caption{
    Representative WorldOlympiad case studies detected by the benchmark.
    The upper examples show high-quality generations that preserve the intended
    physical behavior, scene structure, or interaction state, while the lower
    examples show typical failure cases with visible rule violations,
    geometric inconsistency, or interaction drift.
}
\label{fig:failure-case-study}
\end{figure}

\subsection{Qualitative Case Studies and Failure Modes}
\label{sec:case-study-failure-modes}

Quantitative scores are paired with qualitative cases because a leaderboard
alone cannot explain model failures. As shown in
Figure~\ref{fig:failure-case-study}, WorldOlympiad reveals three recurring
failure modes. Physical metrics identify implausible dynamics, such as objects
moving against gravity, deforming without contact, or changing state abruptly.
Geometry metrics expose videos that look reasonable in the original view but
fail under 3D reconstruction, meta-view rendering, or camera-trajectory
comparison. Interaction metrics capture rollouts that follow isolated captions
but reset state, lose objects, or break action continuity across chunks.
Additional qualitative examples and discussion are provided in
Appendix~\ref{app:case-study}.

\FloatBarrier

\subsection{Human Preference Alignment}
\label{sec:human-preference-alignment}

To examine whether the WorldOlympiad automatic evaluator is consistent with
human preference, we conduct a controlled alignment study on the evaluated world
models. Since long-video world modeling requires more than visual realism alone,
human annotators compare generated videos from multiple complementary aspects,
including overall perceived quality, physical plausibility, temporal coherence,
and interaction fidelity. These criteria are designed to reflect the key
capabilities targeted by WorldOlympiad and provide a human-centered reference for
evaluating model behavior in downstream scenarios.

We aggregate the annotations into a pairwise human preference score
$S^{\mathrm{human}}$, where higher values indicate stronger human preference.
Table~\ref{tab:human-alignment} compares the resulting human ranking with the
WorldOlympiad automatic ranking over the eight annotated models. The two rankings
are highly consistent, with a Spearman correlation coefficient of
$\rho = 0.95$. This strong agreement suggests that WorldOlympiad's automatic
evaluation captures model-level quality differences that are also perceived by
human annotators. Meanwhile, unlike human evaluation, the automatic evaluator can
be applied at a larger scale and provides more fine-grained diagnostic scores
across physical, geometric, and interaction-related dimensions. These results
indicate that WorldOlympiad offers a scalable yet human-aligned evaluation
protocol for long-video world models. Additional annotation and aggregation
details are provided in Appendix~\ref{app:human-preference-details}.

\begin{table}[htbp]
    \centering
    \small
    \caption{
        Alignment between human preference rankings and WorldOlympiad automatic
        rankings. \(S^{\mathrm{human}}\) denotes the pairwise human preference
        score, and \(S^{\mathrm{auto}}\) denotes the WorldOlympiad \textbf{All}
        score. Rank gap is computed as human rank minus automatic rank.
    }
    \label{tab:human-alignment}
    \setlength{\tabcolsep}{5pt}
    \renewcommand{\arraystretch}{1.15}
    \begin{adjustbox}{max width=\linewidth}
    \begin{tabular}{l@{\hspace{16pt}}l@{\hspace{12pt}}c@{\hspace{10pt}}c@{\hspace{10pt}}c@{\hspace{8pt}}c@{\hspace{8pt}}c}
        \toprule
        \textbf{Category} & \textbf{Model} & \(\boldsymbol{S^{\mathrm{human}}}\) & \(\boldsymbol{S^{\mathrm{auto}}}\) & \textbf{Human Rank} & \textbf{Auto Rank} & \textbf{Rank Gap} \\
        \midrule
        Gaming World Model & LingBot-World & 0.721 & 0.683 & 1 & 1 & 0 \\
        Robotics World Model & Cosmos-Predict-2.5 & 0.648 & 0.671 & 2 & 2 & 0 \\
        General World Model & Rolling Forcing & 0.579 & 0.610 & 3 & 3 & 0 \\
        General World Model & LongLive & 0.532 & 0.584 & 4 & 5 & -1 \\
        General World Model & Yume-1.5 & 0.491 & 0.604 & 5 & 4 & 1 \\
        General World Model & Hunyuan-WorldPlay & 0.423 & 0.477 & 6 & 6 & 0 \\
        Gaming World Model & Matrix-Game 2.0 & 0.309 & 0.231 & 7 & 8 & -1 \\
        Robotics World Model & WoW & 0.271 & 0.434 & 8 & 7 & 1 \\
        \bottomrule
    \end{tabular}
    \end{adjustbox}
\end{table}

\FloatBarrier

\section{Conclusion}
We presented WorldOlympiad, a benchmark for evaluating video world models beyond surface-level visual quality, measuring three core capabilities: physical faithfulness, geometric consistency, and interaction fidelity. WorldOlympiad combines rule-based physical judging, 3D reconstruction-based geometry diagnostics, and chunk-level plus long-range interaction evaluation, providing a unified protocol for diagnosing whether generated videos behave as reliable world simulations. Experiments across gaming-centric, robotics-centric, and general world-model pipelines reveal that current models remain far from reliable world simulators: even strong models fail on physical rules, 3D structure, or long-horizon state preservation, exposing important gaps between perceptually plausible generation and controllable world modeling.
\paragraph{\textbf{Future work.}}
Future work will extend WorldOlympiad to study how different memory mechanisms affect long-horizon consistency and interactive controllability. Although many recent pipelines introduce memory modules to improve long-video generation, their varying model scales, training data, and architectural designs make it difficult to isolate whether performance gains stem from the memory mechanism itself or from confounding factors. We therefore aim to build a controlled evaluation environment that disentangles memory design from other variables. Relevant designs include KV-cache reuse, explicit 3D scene memory, linear attention, and hybrid temporal-spatial mechanisms. By comparing these under shared data, comparable model capacity, and a unified protocol, future analysis can more clearly reveal which memory forms best support physical consistency, geometric stability, and reliable long-horizon interaction.

\clearpage
\bibliographystyle{plain}
\bibliography{paper}

\clearpage
\appendix

\section{WorldOlympiad Judge Prompt Templates}
\label{app:prompt-templates}

The prompt templates below cover dynamic-object extraction, physical
consistency, interaction quality, and 3D reconstruction quality.

\begin{table}[!htbp]
\centering
\small
\caption{Judge-related prompt families used by WorldOlympiad.}
\label{tab:judge-prompt-scope}
\setlength{\tabcolsep}{5pt}
\renewcommand{\arraystretch}{1.12}
\begin{tabularx}{\linewidth}{p{2.5cm}p{3.1cm}X}
\toprule
\textbf{Component} & \textbf{Prompt family} & \textbf{Role in evaluation} \\
\midrule
Physical & Relevance and compliance judges & Select applicable physical rules and judge whether the generated video follows them against the reference. \\
Interaction & Chunk, transition, and global judges & Score local caption following, boundary smoothness, and long-range consistency. \\
3D & Static-scene rewrite and 3D MLLM scorers & Remove dynamic actors from the judging target and score Gaussian-splat reconstruction quality. \\
Preprocessing & Dynamic-object extraction & Select moving or deforming foreground actors for SAM-based masking and diagnostic videos. \\
\bottomrule
\end{tabularx}
\end{table}

\subsection{Dynamic-Object Extraction Prompt}
\label{app:sam-prompt}

Before physical and 3D scoring, WorldOlympiad uses a MLLM prompt to identify the
primary dynamic or deforming objects for SAM-based visualization, masking, and
background completion.

\begin{tcolorbox}[promptbox,title={Dynamic-object extraction}]
\small
\textbf{System role:} The model acts as an expert at spotting only the primary
physical actors that visibly move or deform in a video.\\
\textbf{Selection rules:} Return the fewest distinct moving or deforming
objects, with a maximum of three. Prefer dominant moving foreground objects and
omit secondary or uncertain objects. Do not include static background, scenery,
floors, tables, walls, tools, supports, containers, or objects that are merely
visible. If an object does not visibly change position, orientation, or shape,
do not return it. Merge duplicates and synonyms into one concise noun of one to
three words.\\
\textbf{User query:} Watch the video and list only the main objects that
visibly move or deform. The query includes \texttt{\{video\_prompt\}} as the
video description.\\
\textbf{Output format:} Return only a JSON array with one to three concise
nouns, such as \texttt{["person", "ball"]}. No explanations are allowed.
\end{tcolorbox}

\subsection{Physical Judge Prompts}
\label{app:physical-judge-prompts}

The physical pipeline first runs a relevance judge on the reference video to
determine which physical rules are applicable. It then runs a compliance judge
that compares the generated candidate against the reference.

\begin{tcolorbox}[promptbox,title={Physical batch relevance judge}]
\small
\textbf{System role:} \texttt{PhysicsFilterBatch}, an expert video physics
relevance evaluator.\\
\textbf{Input:} One reference or ground-truth video, its textual prompt, and a
list of physics questions. Each question contains a
\texttt{question\_id}, dimension, rule text, and success condition.\\
\textbf{Decision rule:} For every question, decide whether the reference video
contains enough visual evidence to judge the physical rule. Use
\texttt{related=true} when the rule can be judged from visible objects,
materials, contacts, state changes, motion, support, heat or cold cues,
liquid/gas/solid behavior, deformation, color or material behavior, burning,
dissolving, or other relevant physical evidence. Use
\texttt{related=false} only when the rule truly cannot be evaluated from the
shown scene.\\
\textbf{User query:} The prompt provides \texttt{\{video\_prompt\}} and
\texttt{\{question\_list\_json\}}, and asks the judge to evaluate relevance for
every \texttt{question\_id}.\\
\textbf{Output format:} Return strict JSON with one result per input question:
\texttt{question\_id}, \texttt{related}, \texttt{confidence}, and a short
evidence-based \texttt{reason}.
\end{tcolorbox}

\begin{tcolorbox}[promptbox,title={Physical batch compliance judge}]
\small
\textbf{System role:} \texttt{PhysicsJudgeBatch}, an expert video physics
evaluator.\\
\textbf{Input:} Two videos are provided: the generated candidate video to judge
and the ground-truth reference video, which supplies context for the intended
event, scene, timing, and physical evidence. The prompt also provides
\texttt{\{video\_prompt\}} and the related subset of
\texttt{\{question\_list\_json\}}.\\
\textbf{Decision rule:} For each related physics question, judge whether the
generated candidate follows the physical rule and remains consistent with the
reference. The judge does not require frame-exact matching, but it requires
plausible physics, object and material identity consistency, correct temporal
order, and visible evidence.\\
\textbf{Output format:} Return strict JSON with one result per input question:
\texttt{question\_id}, \texttt{compliant}, \texttt{confidence}, a 3--5 sentence
\texttt{explanation} with specific visual evidence, and short concrete
\texttt{observations}.
\end{tcolorbox}

\paragraph{Physical question context.}
The \texttt{question\_list\_json} variable contains rule identifiers,
dimensions, questions, and success conditions. These rules cover mechanics
(\texttt{gravity}, \texttt{buoyancy},
\texttt{compression}, \texttt{impact}), thermodynamics (\texttt{melting},
\texttt{sublimation}, \texttt{vaporization}, \texttt{condensation},
\texttt{deposition}, \texttt{freezing}), and material behavior
(\texttt{color\_mixing}, \texttt{solubility}, \texttt{hardness},
\texttt{combustibility}).

\begin{tcolorbox}[promptbox,title={Mechanics rule contexts}]
\small
\textbf{\texttt{gravity}}\\
\texttt{rule}: Do free-moving objects move downward consistently with gravity?\\
\texttt{expected\_behavior}: First judge whether objects are unsupported,
airborne, falling, jumping, driving over uneven terrain, or staying grounded on a
support surface. Unsupported objects should fall or arc downward under gravity;
supported ground vehicles, people, and objects should remain in plausible contact
with the ground unless a visible jump, ramp, collision, or lift explains vertical
motion. Penalize floating, sinking through the ground, sudden vertical pops, or
hovering without support.

\textbf{\texttt{buoyancy}}\\
\texttt{rule}: Do objects on or in a fluid behave consistently with buoyancy,
such that floating items stay near the surface and sinking items submerge?\\
\texttt{expected\_behavior}: Floating objects should remain on or near the
surface; dense objects should descend.

\textbf{\texttt{compression}}\\
\texttt{rule}: When objects or support surfaces are stressed, loaded, squeezed,
or pressed, do they deform or remain rigid in a plausible manner?\\
\texttt{expected\_behavior}: Cans may dent when crushed, soft materials should
compress smoothly under load, and rigid vehicles or metal bodies should mostly
keep their shape unless there is collision or heavy force. In robot manipulation,
vehicle, racing, or navigation scenes, grasping, pinching, pressing, tire-ground
contact, suspension loading, soil displacement, body rigidity, or lack of
impossible warping can make stress and deformation relevant. Deformation should
start only after visible squeezing, support, load, contact, or other applied
stress, and the same object, person, or vehicle should stay visually consistent
instead of morphing into a different shape or identity.

\textbf{\texttt{impact}}\\
\texttt{rule}: Do contact, traction, collisions, impacts, and momentum changes
produce reasonable motion transitions?\\
\texttt{expected\_behavior}: Look for momentum transfer, bouncing, shattering,
resting poses, ground contact, tire or foot contact, traction, braking, turning,
or acceleration that matches visible forces and contacts. In robot manipulation,
placing or releasing an object into contact also counts as an impact or contact
event. In vehicle, racing, sports, or navigation scenes, tire-ground contact,
acceleration from rest, sliding, dust kick-up, braking, steering, direction
changes, speed changes, near-collisions, or collisions are relevant. Abrupt
deformation, direction change, or speed change should happen after visible
contact, impact, or control input rather than before.
\end{tcolorbox}

\begin{tcolorbox}[promptbox,title={Thermodynamics rule contexts}]
\small
\textbf{\texttt{melting}}\\
\texttt{description}: A solid substance should gradually transition to liquid
when heated above its melting point.\\
\texttt{expected\_behavior}: The solid should decrease in size or volume as it
transforms into liquid, and the process should be gradual and continuous.\\
\texttt{violations}: Solid increasing in size, liquid turning into solid,
instantaneous disappearance without liquid formation, solid remaining unchanged
despite heat, or no liquid formation.

\textbf{\texttt{sublimation}}\\
\texttt{description}: A solid should directly transition to gas without passing
through the liquid phase.\\
\texttt{expected\_behavior}: The solid should rapidly decrease or disappear
while gas or vapor appears around it, with no visible liquid phase.\\
\texttt{violations}: Solid melting to liquid first, gas condensing to solid,
solid remaining stable, or liquid forming as an intermediate phase.

\textbf{\texttt{vaporization}}\\
\texttt{description}: A liquid should transition to gas when heated or over time.\\
\texttt{expected\_behavior}: The liquid should decrease in volume or disappear
over time; bubbles, boiling, evaporation, or visible gas or vapor may appear.\\
\texttt{violations}: Liquid increasing in volume, gas condensing to liquid,
liquid remaining constant despite heat or time, or liquid freezing.

\textbf{\texttt{condensation}}\\
\texttt{description}: A gas should transition to liquid when cooled below its
condensation point.\\
\texttt{expected\_behavior}: Liquid should form and increase in volume; small
droplets may appear and merge, and gas may become less visible as it condenses.\\
\texttt{violations}: Liquid evaporating, gas remaining as gas, liquid freezing
directly to solid, or no liquid formation.

\textbf{\texttt{deposition}}\\
\texttt{description}: A gas should directly transition to solid without passing
through the liquid phase.\\
\texttt{expected\_behavior}: Solid should form and increase in size or volume
directly from gas, often as crystals or frost-like structures, with no visible
liquid intermediate.\\
\texttt{violations}: Gas condensing to liquid first, solid melting, liquid
freezing to solid, or no solid formation.

\textbf{\texttt{freezing}}\\
\texttt{description}: A liquid should transition to solid when cooled below its
freezing point.\\
\texttt{expected\_behavior}: The liquid should solidify and may expand slightly
or contract; the surface may become rigid and the shape should become more
defined and regular.\\
\texttt{violations}: Liquid remaining liquid, solid melting, liquid evaporating,
or no solidification occurring.
\end{tcolorbox}

\begin{tcolorbox}[promptbox,title={Material rule contexts}]
\small
\textbf{\texttt{color\_mixing}}\\
\texttt{question}: When different colored liquids or paints mix, do they produce
the correct resulting color?\\
\texttt{success\_condition}: Colored liquids, paints, powders, smoke, dye,
pigments, or other visibly colored substances should blend into plausible
resulting colors when they contact and mix. Red and yellow should produce
orange, blue and yellow should produce green, and red and blue should produce
purple. If colored objects merely pass near each other without material transfer
or blending, they should not change color.

\textbf{\texttt{solubility}}\\
\texttt{question}: Do soluble materials such as sugar or salt dissolve properly
when placed in water or other solvents?\\
\texttt{success\_condition}: Soluble or dispersible substances such as sugar,
salt, powder, dye, tablets, or granular material should gradually disperse,
dissolve, fade, or become suspended or invisible in a liquid solvent. Insoluble
solids should remain visible or settle.

\textbf{\texttt{hardness}}\\
\texttt{question}: Do materials with different hardness levels behave correctly
when grasped, pressed, cut, folded, or broken?\\
\texttt{success\_condition}: Soft materials such as paper, cloth, foam, food,
plants, soil, loose dirt, or powder should bend, fold, tear, compress, scatter,
or deform when appropriate. Hard materials such as metal, stone, glass, rigid
vehicle bodies, tools, or containers should resist deformation unless force or
collision is strong enough. In robot, human, vehicle, racing, sports, or
navigation scenes, grasping, pinching, pressing, stepping, tire-ground contact,
placing, sliding, collision, or load-bearing can reveal rigidity versus softness.
Shape change should follow visible contact or applied force, and the acted-on
object, person, vehicle, or material should remain visually consistent.

\textbf{\texttt{combustibility}}\\
\texttt{question}: Do flammable materials burn correctly, producing fire, smoke,
or char?\\
\texttt{success\_condition}: Wood, paper, and fabric should ignite and produce
flames or smoke; non-flammable materials should not.
\end{tcolorbox}

\subsection{Interaction Judge Prompts}
\label{app:interaction-judge-prompts}

The interaction pipeline evaluates chunk-generated long videos at three
levels: individual chunks, adjacent chunk transitions, and the stitched full
video.

\begin{tcolorbox}[promptbox,title={Interaction judge shared instruction}]
\small
\textbf{System role:} A strict evaluator for chunk-generated interactive
videos.\\
\textbf{Judgment scope:} Judge whether each generated chunk follows its
intended action and caption, whether adjacent chunks transition smoothly, and
whether the full stitched video remains globally consistent. The reference video
is used only as context for scene, style, camera, and intended interaction, not
as a requirement for frame-exact matching.\\
\textbf{Output rule:} Always return strict JSON only. All scores must be
numbers from 0 to 5.
\end{tcolorbox}

\begin{tcolorbox}[promptbox,title={Chunk-level interaction judge}]
\small
\textbf{Input metadata:} \texttt{\{chunk\_index\}},
\texttt{\{source\_interval\}}, generated interval
\texttt{[\{generated\_start\_sec\}, \{generated\_end\_sec\})},
\texttt{\{action\}}, and \texttt{\{caption\}}.\\
\textbf{Primary evidence:} Use the generated chunk frames as the primary
evidence. Reference-video frames are used only for context about the intended
scene and style.\\
\textbf{Scoring criteria:} Score \texttt{visual\_quality} by clarity, realism,
temporal stability, color and lighting consistency, and lack of artifacts. Score
\texttt{text\_alignment} by whether the visible content matches the intended
action and caption.\\
\textbf{Output format:} Return strict JSON with \texttt{chunk\_index},
\texttt{visual\_quality}, \texttt{text\_alignment}, \texttt{overall}, and a
short evidence-based \texttt{reason}.
\end{tcolorbox}

\begin{tcolorbox}[promptbox,title={Transition-level interaction judge}]
\small
\textbf{Input metadata:} The previous and next chunk indices, generated
intervals, actions, and captions.\\
\textbf{Decision rule:} Judge whether the transition is smooth and continuous.
Scene, lighting, style, and subject identity should remain coherent. Motion
trajectory and camera movement should evolve naturally. Penalize abrupt jumps,
object identity resets, impossible camera jumps, and visible stitching
artifacts.\\
\textbf{Output format:} Return strict JSON with \texttt{from\_chunk\_index},
\texttt{to\_chunk\_index}, \texttt{transition\_smoothness}, \texttt{overall},
and a short evidence-based \texttt{reason}.
\end{tcolorbox}

\begin{tcolorbox}[promptbox,title={Global interaction judge}]
\small
\textbf{Input:} The whole stitched generated video and
\texttt{\{prompt\_summary\_json\}}, which lists each chunk index, action, and
caption.\\
\textbf{Primary evidence:} Use generated full-video frames as the primary
evidence and reference frames only as context.\\
\textbf{Scoring criteria:} Judge whether subject, character, and object
identity remain stable across the video; whether scene style, visual tone,
lighting, and camera behavior remain coherent; and whether global semantics
align with the combined intent of all chunk prompts.\\
\textbf{Output format:} Return strict JSON with
\texttt{long\_range\_consistency}, \texttt{global\_text\_alignment},
\texttt{overall}, and a short evidence-based \texttt{reason}.
\end{tcolorbox}

\subsection{3D Judge Prompts}
\label{app:3d-judge-prompts}

The 3D pipeline first rewrites the original generation prompt into a static
scene prompt, because dynamic foreground actors are masked and video-inpainted
before Depth Anything 3 reconstruction and Gaussian-Splat rendering. The MLLM
then scores the Gaussian-Splat video and a meta-view image. The camera
trajectory score \(S_{\mathrm{traj}}\) is computed from DA3 camera motion
similarity.

\begin{tcolorbox}[promptbox,title={Static-scene prompt rewrite}]
\small
\textbf{System role:} Rewrite video generation prompts for static 3D
reconstruction evaluation. The reconstruction input has dynamic foreground
actors masked and video-inpainted before DA3 reconstruction.\\
\textbf{Rewrite requirements:} Keep static background, environment, layout,
materials, lighting, weather, terrain, large structures, and camera/view
behavior if present. Remove or explicitly ignore dynamic foreground actors and
actions, including people, animals, vehicles, limbs, clothing motion, object
manipulation, running, turning, and other moving subjects. If the original
prompt has no camera information, set \texttt{camera\_behavior} exactly to
\texttt{"camera is static"}. Mention that dynamic foreground actors may be
absent because they were masked and video-inpainted before reconstruction.\\
\textbf{Input placeholder:} \texttt{\{prompt\}}.\\
\textbf{Output format:} Return strict JSON with exactly
\texttt{static\_scene\_description}, \texttt{camera\_behavior}, and
\texttt{ignore\_for\_3d}.
\end{tcolorbox}

\begin{tcolorbox}[promptbox,title={Gaussian-Splat video 3D judge}]
\small
\textbf{System role:} A calibrated 3D reconstruction judge that follows the
scoring rubric and returns strict JSON only.\\
\textbf{Input:} Several sampled frames from a Gaussian-Splat render
reconstructed from a source video after dynamic foreground regions were masked
and video-inpainted, together with the static-scene prompt
\texttt{\{prompt\}}.\\
\textbf{Robustness instruction:} Be robust to normal Gaussian-Splat artifacts,
video-inpainting artifacts, blur, rain/fog/low-light appearance, and moderate
texture noise. Dynamic foreground actors and actions may be absent or
incomplete because they were intentionally removed before reconstruction; do not
penalize their absence.\\
\textbf{Scoring focus:} Judge whether the reconstructed static background
geometry is coherent across views or time, whether static structures and scene
layout are spatially plausible, whether the render preserves recognizable
camera-consistent organization, whether it is faithful to the static-scene
description and expected camera behavior, and whether artifacts dominate the
render. If the prompt says the camera is static, do not require camera motion or
parallax.\\
\textbf{Calibration:} Scores in \(0.8\)--\(1.0\) indicate a clear coherent 3D
scene; \(0.6\)--\(0.8\) indicate a recognizable mostly coherent scene with
noticeable artifacts; \(0.4\)--\(0.6\) indicate partial recognition with
significant artifacts; \(0.2\)--\(0.4\) indicate mostly failed reconstruction;
and \(0.0\)--\(0.2\) indicate an unusable render. Do not assign a very low
score solely because the render is blurry or noisy; if the static layout is
recognizable, the score should usually be at least \(0.5\).\\
\textbf{Output format:} Return strict JSON with \texttt{score} in \([0,1]\) and
a short \texttt{reason}.
\end{tcolorbox}

\begin{tcolorbox}[promptbox,title={Meta-view image 3D judge}]
\small
\textbf{System role:} A calibrated 3D reconstruction judge that follows the
scoring rubric and returns strict JSON only.\\
\textbf{Input:} A single meta-view image rendered from a 3D reconstruction of a
source video after dynamic foreground regions were masked and video-inpainted,
together with the static-scene prompt \texttt{\{prompt\}}.\\
\textbf{Robustness instruction:} Be robust to normal Gaussian-Splat artifacts,
video-inpainting artifacts, blur, rain/fog/low-light appearance, and moderate
texture noise. Dynamic foreground actors and actions may be absent or
incomplete because they were intentionally removed before reconstruction; do not
penalize their absence.\\
\textbf{Scoring focus:} Judge whether the static background layout is
recognizable, geometrically plausible, structurally coherent, and not
catastrophically flat, floating, duplicated, or collapsed. If the prompt says
the camera is static, do not require camera motion or parallax.\\
\textbf{Calibration:} Scores in \(0.8\)--\(1.0\) indicate a clear coherent
static scene; \(0.6\)--\(0.8\) indicate a recognizable mostly coherent scene
with noticeable artifacts; \(0.4\)--\(0.6\) indicate a partially recognizable
scene; \(0.2\)--\(0.4\) indicate a mostly failed reconstruction; and
\(0.0\)--\(0.2\) indicate an unusable image. Do not assign a very low score
solely because the meta-view is blurry or noisy; if the static scene layout is
recognizable, the score should usually be at least \(0.5\).\\
\textbf{Output format:} Return strict JSON with \texttt{score} in \([0,1]\) and
a short \texttt{reason}.
\end{tcolorbox}

\section{Detailed Results}
\label{app:detailed-results}

This section reports domain-wise scores, physical pass rates, interaction
diagnostics, geometry diagnostics, and model-level submetrics.

\subsection{Domain-wise Results}
\label{app:detailed-domain-results}

Table~\ref{tab:detailed-domain-results} reports the detailed scores on the
same-scene subset, grouped by evaluation domain. The table includes physical
faithfulness, 3D consistency, CLIP-augmented interaction fidelity, raw and
calibrated CLIP semantic alignment, and the overall score. The overall score
is computed as the equal-weight average of physical faithfulness, 3D
consistency, and interaction fidelity.

\begin{table}[!htbp]
\centering
\small
\caption{Detailed WorldOlympiad scores on the same-scene subset across gaming, robotics, and general domains. \textbf{All} is the equal-weight average of \textbf{Physical}, \textbf{3D Cons.}, and \textbf{Interact.}}
\label{tab:detailed-domain-results}
\setlength{\tabcolsep}{4pt}
\renewcommand{\arraystretch}{1.12}
\begin{tabularx}{\linewidth}{p{1.6cm}>{\raggedright\arraybackslash}Xcccccc}
\toprule
\textbf{Domain} & \textbf{Pipeline} & \textbf{Physical} & \textbf{3D Cons.} & \textbf{Interact.} & \textbf{CLIP Raw} & \textbf{CLIP Aux.} & \textbf{All} \\
\midrule
\multirow{8}{*}{Gaming}
& Matrix-Game 2.0 & 0.332 & 0.189 & 0.111 & 0.230 & 0.150 & 0.211 \\
& LingBot-World & 0.884 & 0.366 & 0.778 & 0.315 & 0.575 & 0.676 \\
& Cosmos-Predict-2.5 & 0.867 & 0.361 & 0.679 & 0.306 & 0.530 & 0.636 \\
& WoW & 0.633 & 0.223 & 0.249 & 0.247 & 0.235 & 0.368 \\
& Rolling Forcing & 0.853 & 0.289 & 0.675 & 0.332 & 0.660 & 0.606 \\
& LongLive & 0.851 & 0.292 & 0.554 & 0.322 & 0.610 & 0.566 \\
& Yume-1.5 & 0.813 & 0.352 & 0.659 & 0.291 & 0.455 & 0.608 \\
& Hunyuan-WorldPlay & 0.852 & 0.348 & 0.471 & 0.296 & 0.480 & 0.557 \\
\midrule
\multirow{8}{*}{Robotics}
& Matrix-Game 2.0 & 0.364 & 0.338 & 0.139 & 0.252 & 0.260 & 0.280 \\
& LingBot-World & 0.949 & 0.393 & 0.710 & 0.314 & 0.570 & 0.684 \\
& Cosmos-Predict-2.5 & 0.937 & 0.479 & 0.721 & 0.321 & 0.605 & 0.712 \\
& WoW & 0.787 & 0.272 & 0.447 & 0.288 & 0.440 & 0.502 \\
& Rolling Forcing & 0.870 & 0.389 & 0.566 & 0.329 & 0.645 & 0.608 \\
& LongLive & 0.857 & 0.472 & 0.470 & 0.327 & 0.635 & 0.600 \\
& Yume-1.5 & 0.851 & 0.288 & 0.624 & 0.312 & 0.560 & 0.588 \\
& Hunyuan-WorldPlay & 0.630 & 0.600 & 0.262 & 0.309 & 0.545 & 0.497 \\
\midrule
\multirow{8}{*}{General}
& Matrix-Game 2.0 & 0.216 & 0.220 & 0.067 & 0.222 & 0.110 & 0.168 \\
& LingBot-World & 0.963 & 0.335 & 0.767 & 0.311 & 0.555 & 0.688 \\
& Cosmos-Predict-2.5 & 0.939 & 0.317 & 0.736 & 0.313 & 0.565 & 0.664 \\
& WoW & 0.692 & 0.251 & 0.302 & 0.256 & 0.280 & 0.415 \\
& Rolling Forcing & 0.933 & 0.285 & 0.657 & 0.314 & 0.570 & 0.625 \\
& LongLive & 0.909 & 0.290 & 0.579 & 0.315 & 0.575 & 0.593 \\
& Yume-1.5 & 0.925 & 0.302 & 0.694 & 0.302 & 0.510 & 0.640 \\
& Hunyuan-WorldPlay & 0.389 & 0.219 & 0.097 & 0.235 & 0.175 & 0.235 \\
\bottomrule
\end{tabularx}
\end{table}

\FloatBarrier

\subsection{Fine-grained Physical Results}
\label{app:fine-grained-physical-results}

Table~\ref{tab:physical-dimension-results} reports physical pass rates
aggregated by physical dimension. Table~\ref{tab:physical-question-results}
further breaks these scores down into individual physical questions.

\begin{table}[!htbp]
\centering
\small
\caption{Physical dimension pass rates on the same-scene subset.}
\label{tab:physical-dimension-results}
\setlength{\tabcolsep}{4pt}
\renewcommand{\arraystretch}{1.12}
\begin{tabularx}{\linewidth}{p{1.6cm}>{\raggedright\arraybackslash}Xcccc}
\toprule
\textbf{Domain} & \textbf{Pipeline} & \textbf{Overall} & \textbf{Mechanics} & \textbf{Thermodynamics} & \textbf{Material} \\
\midrule
\multirow{8}{*}{Gaming}
& Matrix-Game 2.0 & 0.332 & 0.433 & 0.172 & 0.184 \\
& LingBot-World & 0.884 & 0.983 & 0.450 & 0.969 \\
& Cosmos-Predict-2.5 & 0.867 & 0.951 & 0.418 & 0.884 \\
& WoW & 0.633 & 0.806 & 0.226 & 0.446 \\
& Rolling Forcing & 0.853 & 0.941 & 0.418 & 0.854 \\
& LongLive & 0.851 & 0.941 & 0.377 & 0.865 \\
& Yume-1.5 & 0.813 & 0.942 & 0.365 & 0.902 \\
& Hunyuan-WorldPlay & 0.852 & 0.944 & 0.426 & 0.843 \\
\midrule
\multirow{8}{*}{Robotics}
& Matrix-Game 2.0 & 0.364 & 0.366 & 0.000 & 0.372 \\
& LingBot-World & 0.949 & 0.961 & 0.000 & 0.957 \\
& Cosmos-Predict-2.5 & 0.937 & 0.939 & 0.000 & 0.968 \\
& WoW & 0.787 & 0.798 & 0.111 & 0.788 \\
& Rolling Forcing & 0.870 & 0.857 & 0.000 & 0.935 \\
& LongLive & 0.857 & 0.864 & 0.000 & 0.869 \\
& Yume-1.5 & 0.851 & 0.857 & 0.000 & 0.872 \\
& Hunyuan-WorldPlay & 0.630 & 0.577 & 0.000 & 0.810 \\
\midrule
\multirow{8}{*}{General}
& Matrix-Game 2.0 & 0.216 & 0.246 & 0.097 & 0.036 \\
& LingBot-World & 0.963 & 1.000 & 0.519 & 1.000 \\
& Cosmos-Predict-2.5 & 0.939 & 0.977 & 0.613 & 0.875 \\
& WoW & 0.692 & 0.743 & 0.300 & 0.562 \\
& Rolling Forcing & 0.933 & 0.968 & 0.581 & 0.938 \\
& LongLive & 0.909 & 0.952 & 0.581 & 0.812 \\
& Yume-1.5 & 0.925 & 0.979 & 0.370 & 0.906 \\
& Hunyuan-WorldPlay & 0.389 & 0.430 & 0.323 & 0.036 \\
\bottomrule
\end{tabularx}
\end{table}

\begin{table}[!htbp]
\centering
\scriptsize
\caption{Physical question pass rates on the same-scene subset.}
\label{tab:physical-question-results}
\setlength{\tabcolsep}{2.5pt}
\renewcommand{\arraystretch}{1.08}
\begin{tabularx}{\linewidth}{p{1.2cm}>{\raggedright\arraybackslash}Xcccccccccccccc}
\toprule
\textbf{Domain} & \textbf{Pipeline} & \textbf{Grav.} & \textbf{Buoy.} & \textbf{Comp.} & \textbf{Impact} & \textbf{Melt} & \textbf{Sub.} & \textbf{Vap.} & \textbf{Cond.} & \textbf{Dep.} & \textbf{Freez.} & \textbf{Color} & \textbf{Sol.} & \textbf{Hard.} & \textbf{Comb.} \\
\midrule
\multirow{8}{*}{Gaming}
& Matrix-Game 2.0 & 0.494 & 0.479 & 0.324 & 0.247 & 0.214 & 0.000 & 0.146 & 0.179 & 0.167 & 0.231 & -- & -- & 0.168 & 0.222 \\
& LingBot-World & 0.986 & 0.944 & 1.000 & 1.000 & 0.429 & -- & 0.462 & 0.417 & 0.500 & 0.500 & -- & -- & 0.976 & 0.957 \\
& Cosmos-Predict-2.5 & 0.958 & 0.986 & 0.986 & 0.868 & 0.357 & 0.000 & 0.292 & 0.513 & 0.833 & 0.538 & -- & -- & 0.901 & 0.843 \\
& WoW & 0.850 & 0.944 & 0.774 & 0.540 & 0.200 & 0.000 & 0.161 & 0.296 & 0.200 & 0.333 & -- & -- & 0.486 & 0.355 \\
& Rolling Forcing & 0.949 & 0.987 & 0.947 & 0.871 & 0.357 & 0.500 & 0.298 & 0.475 & 0.667 & 0.615 & -- & -- & 0.854 & 0.854 \\
& LongLive & 0.955 & 0.986 & 0.946 & 0.846 & 0.429 & 0.000 & 0.292 & 0.436 & 0.500 & 0.462 & -- & -- & 0.875 & 0.843 \\
& Yume-1.5 & 0.955 & 1.000 & 0.900 & 0.786 & 0.600 & 0.000 & 0.286 & 0.333 & 0.500 & 0.600 & -- & -- & 0.902 & 0.900 \\
& Hunyuan-WorldPlay & 0.957 & 0.986 & 0.959 & 0.844 & 0.500 & 0.500 & 0.271 & 0.513 & 0.667 & 0.538 & -- & -- & 0.870 & 0.780 \\
\midrule
\multirow{8}{*}{Robotics}
& Matrix-Game 2.0 & 0.427 & 0.467 & 0.290 & 0.239 & -- & -- & 0.000 & -- & -- & -- & -- & 0.000 & 0.374 & -- \\
& LingBot-World & 0.965 & 1.000 & 0.985 & 0.935 & -- & -- & 0.000 & -- & -- & -- & 1.000 & 0.500 & 0.962 & -- \\
& Cosmos-Predict-2.5 & 0.945 & 1.000 & 1.000 & 0.894 & -- & -- & 0.000 & -- & -- & -- & 1.000 & 0.500 & 0.972 & -- \\
& WoW & 0.840 & 0.929 & 0.845 & 0.662 & -- & -- & 0.111 & -- & -- & -- & 0.000 & 0.000 & 0.800 & -- \\
& Rolling Forcing & 0.889 & 1.000 & 0.889 & 0.766 & -- & -- & 0.000 & -- & -- & -- & -- & 0.000 & 0.941 & -- \\
& LongLive & 0.879 & 0.933 & 0.938 & 0.791 & -- & -- & 0.000 & -- & -- & -- & -- & 0.000 & 0.873 & -- \\
& Yume-1.5 & 0.874 & 1.000 & 0.938 & 0.765 & -- & -- & 0.000 & -- & -- & -- & 0.000 & 0.000 & 0.885 & -- \\
& Hunyuan-WorldPlay & 0.644 & 0.812 & 0.615 & 0.373 & -- & -- & 0.000 & -- & -- & -- & 0.000 & 0.000 & 0.822 & -- \\
\midrule
\multirow{8}{*}{General}
& Matrix-Game 2.0 & 0.310 & 0.267 & 0.111 & 0.130 & 0.125 & -- & 0.000 & 0.000 & 1.000 & 0.000 & -- & -- & 0.037 & 0.000 \\
& LingBot-World & 1.000 & 1.000 & 1.000 & 1.000 & 0.600 & -- & 0.400 & 0.500 & 1.000 & 0.667 & -- & -- & 1.000 & 1.000 \\
& Cosmos-Predict-2.5 & 0.972 & 1.000 & 1.000 & 0.976 & 0.875 & -- & 0.333 & 1.000 & 1.000 & 0.800 & -- & -- & 0.903 & 0.000 \\
& WoW & 0.767 & 0.806 & 0.889 & 0.634 & 0.500 & -- & 0.133 & -- & 0.500 & 0.400 & -- & -- & 0.581 & 0.000 \\
& Rolling Forcing & 0.978 & 0.935 & 1.000 & 0.951 & 0.875 & -- & 0.267 & 1.000 & 1.000 & 0.800 & -- & -- & 0.935 & 1.000 \\
& LongLive & 0.956 & 1.000 & 1.000 & 0.915 & 0.875 & -- & 0.400 & 0.000 & 0.500 & 0.800 & -- & -- & 0.839 & 0.000 \\
& Yume-1.5 & 0.983 & 1.000 & 1.000 & 0.958 & 0.400 & -- & 0.267 & 0.500 & 1.000 & 0.333 & -- & -- & 0.903 & 1.000 \\
& Hunyuan-WorldPlay & 0.494 & 0.467 & 0.500 & 0.260 & 0.500 & -- & 0.067 & 0.000 & 1.000 & 0.600 & -- & -- & 0.037 & 0.000 \\
\bottomrule
\end{tabularx}
\end{table}

\FloatBarrier

\subsection{Fine-grained Interaction Results}
\label{app:fine-grained-interaction-results}

Table~\ref{tab:fine-grained-interaction-results} reports fine-grained
interaction diagnostics. The chunk score measures local caption and action
following, the transition score measures boundary smoothness between adjacent
chunks, and the global score measures long-range consistency over the stitched
video. The raw CLIP score is calibrated into a bounded auxiliary score with
fixed thresholds, and the interaction score corresponds to the aggregate
interaction metric reported in Table~\ref{tab:detailed-domain-results}.

\begin{table}[!htbp]
\centering
\small
\caption{Fine-grained interaction diagnostics on the same-scene subset.}
\label{tab:fine-grained-interaction-results}
\setlength{\tabcolsep}{4pt}
\renewcommand{\arraystretch}{1.12}
\begin{tabularx}{\linewidth}{p{1.5cm}>{\raggedright\arraybackslash}Xcccccccc}
\toprule
\textbf{Domain} & \textbf{Pipeline} & \textbf{Chunk} & \textbf{Trans.} & \textbf{Global} & \textbf{Long Range} & \textbf{Global Text} & \textbf{CLIP Raw} & \textbf{CLIP Aux.} & \textbf{Interact.} \\
\midrule
\multirow{8}{*}{Gaming}
& Matrix-Game 2.0 & 0.135 & 0.074 & 0.087 & 0.087 & 0.087 & 0.230 & 0.150 & 0.111 \\
& LingBot-World & 0.796 & 0.767 & 0.862 & 0.875 & 0.848 & 0.315 & 0.575 & 0.778 \\
& Cosmos-Predict-2.5 & 0.704 & 0.677 & 0.700 & 0.718 & 0.680 & 0.306 & 0.530 & 0.679 \\
& WoW & 0.267 & 0.233 & 0.247 & 0.250 & 0.244 & 0.247 & 0.235 & 0.249 \\
& Rolling Forcing & 0.665 & 0.681 & 0.704 & 0.733 & 0.675 & 0.332 & 0.660 & 0.675 \\
& LongLive & 0.595 & 0.444 & 0.625 & 0.640 & 0.606 & 0.322 & 0.610 & 0.554 \\
& Yume-1.5 & 0.645 & 0.727 & 0.668 & 0.702 & 0.632 & 0.291 & 0.455 & 0.659 \\
& Hunyuan-WorldPlay & 0.483 & 0.458 & 0.440 & 0.464 & 0.415 & 0.296 & 0.480 & 0.471 \\
\midrule
\multirow{8}{*}{Robotics}
& Matrix-Game 2.0 & 0.136 & 0.167 & 0.041 & 0.042 & 0.041 & 0.252 & 0.260 & 0.139 \\
& LingBot-World & 0.670 & 0.714 & 0.881 & 0.893 & 0.869 & 0.314 & 0.570 & 0.710 \\
& Cosmos-Predict-2.5 & 0.682 & 0.707 & 0.896 & 0.908 & 0.885 & 0.321 & 0.605 & 0.721 \\
& WoW & 0.413 & 0.501 & 0.472 & 0.493 & 0.451 & 0.288 & 0.440 & 0.447 \\
& Rolling Forcing & 0.498 & 0.600 & 0.632 & 0.661 & 0.603 & 0.329 & 0.645 & 0.566 \\
& LongLive & 0.484 & 0.288 & 0.587 & 0.619 & 0.556 & 0.327 & 0.635 & 0.470 \\
& Yume-1.5 & 0.553 & 0.715 & 0.694 & 0.722 & 0.667 & 0.312 & 0.560 & 0.624 \\
& Hunyuan-WorldPlay & 0.245 & 0.254 & 0.134 & 0.140 & 0.129 & 0.309 & 0.545 & 0.262 \\
\midrule
\multirow{8}{*}{General}
& Matrix-Game 2.0 & 0.069 & 0.064 & 0.031 & 0.031 & 0.029 & 0.222 & 0.110 & 0.067 \\
& LingBot-World & 0.752 & 0.819 & 0.829 & 0.838 & 0.812 & 0.311 & 0.555 & 0.767 \\
& Cosmos-Predict-2.5 & 0.746 & 0.755 & 0.764 & 0.782 & 0.746 & 0.313 & 0.565 & 0.736 \\
& WoW & 0.314 & 0.294 & 0.292 & 0.293 & 0.285 & 0.256 & 0.280 & 0.302 \\
& Rolling Forcing & 0.620 & 0.727 & 0.661 & 0.684 & 0.629 & 0.314 & 0.570 & 0.657 \\
& LongLive & 0.598 & 0.520 & 0.639 & 0.661 & 0.603 & 0.315 & 0.575 & 0.579 \\
& Yume-1.5 & 0.637 & 0.814 & 0.718 & 0.744 & 0.684 & 0.302 & 0.510 & 0.694 \\
& Hunyuan-WorldPlay & 0.130 & 0.030 & 0.073 & 0.073 & 0.073 & 0.235 & 0.175 & 0.097 \\
\bottomrule
\end{tabularx}
\end{table}

\FloatBarrier

\subsection{Fine-grained Geometry Results}
\label{app:fine-grained-geometry-results}

Table~\ref{tab:fine-grained-geometry-results} reports fine-grained geometry
diagnostics. \(S_{\mathrm{recon}}\) measures the quality of the Gaussian-splat
reconstruction video, \(S_{\mathrm{meta}}\) measures the quality of the rendered
meta-view image, and \(S_{\mathrm{traj}}\) measures camera-trajectory
consistency. The 3D consistency score corresponds to the aggregate geometry
metric reported in Table~\ref{tab:detailed-domain-results}.

\begin{table}[!htbp]
\centering
\small
\caption{Fine-grained geometry diagnostics on the same-scene subset.}
\label{tab:fine-grained-geometry-results}
\setlength{\tabcolsep}{4pt}
\renewcommand{\arraystretch}{1.12}
\begin{tabularx}{\linewidth}{p{1.6cm}>{\raggedright\arraybackslash}Xcccc}
\toprule
\textbf{Domain} & \textbf{Pipeline} & \(\boldsymbol{S_{\mathrm{recon}}}\) & \(\boldsymbol{S_{\mathrm{meta}}}\) & \(\boldsymbol{S_{\mathrm{traj}}}\) & \textbf{3D Cons.} \\
\midrule
\multirow{8}{*}{Gaming}
& Matrix-Game 2.0 & 0.160 & 0.159 & 0.247 & 0.189 \\
& LingBot-World & 0.389 & 0.372 & 0.337 & 0.366 \\
& Cosmos-Predict-2.5 & 0.415 & 0.388 & 0.280 & 0.361 \\
& WoW & 0.232 & 0.205 & 0.231 & 0.223 \\
& Rolling Forcing & 0.324 & 0.292 & 0.250 & 0.289 \\
& LongLive & 0.328 & 0.292 & 0.256 & 0.292 \\
& Yume-1.5 & 0.381 & 0.361 & 0.315 & 0.352 \\
& Hunyuan-WorldPlay & 0.397 & 0.363 & 0.284 & 0.348 \\
\midrule
\multirow{8}{*}{Robotics}
& Matrix-Game 2.0 & 0.283 & 0.298 & 0.432 & 0.338 \\
& LingBot-World & 0.416 & 0.416 & 0.348 & 0.393 \\
& Cosmos-Predict-2.5 & 0.451 & 0.464 & 0.523 & 0.479 \\
& WoW & 0.297 & 0.289 & 0.232 & 0.272 \\
& Rolling Forcing & 0.458 & 0.432 & 0.278 & 0.389 \\
& LongLive & 0.476 & 0.483 & 0.458 & 0.472 \\
& Yume-1.5 & 0.337 & 0.340 & 0.185 & 0.288 \\
& Hunyuan-WorldPlay & 0.574 & 0.566 & 0.660 & 0.600 \\
\midrule
\multirow{8}{*}{General}
& Matrix-Game 2.0 & 0.191 & 0.196 & 0.271 & 0.220 \\
& LingBot-World & 0.373 & 0.319 & 0.312 & 0.335 \\
& Cosmos-Predict-2.5 & 0.341 & 0.322 & 0.288 & 0.317 \\
& WoW & 0.243 & 0.225 & 0.286 & 0.251 \\
& Rolling Forcing & 0.283 & 0.266 & 0.306 & 0.285 \\
& LongLive & 0.289 & 0.280 & 0.300 & 0.290 \\
& Yume-1.5 & 0.318 & 0.306 & 0.282 & 0.302 \\
& Hunyuan-WorldPlay & 0.177 & 0.191 & 0.288 & 0.219 \\
\bottomrule
\end{tabularx}
\end{table}

\FloatBarrier

\subsection{Model-level Fine-grained Results}
\label{app:model-level-fine-grained-results}

Table~\ref{tab:model-level-geometry-results} and
Table~\ref{tab:model-level-interaction-results} aggregate the fine-grained
geometry and interaction diagnostics at the model-category level.

\begin{table}[!htbp]
\centering
\small
\caption{Model-level 3D consistency submetrics.}
\label{tab:model-level-geometry-results}
\setlength{\tabcolsep}{4pt}
\renewcommand{\arraystretch}{1.12}
\begin{tabularx}{\linewidth}{p{2.4cm}>{\raggedright\arraybackslash}Xcccc}
\toprule
\textbf{Category} & \textbf{Model} & \textbf{GS} & \textbf{Meta} & \textbf{Camera Motion} & \textbf{3D Cons.} \\
\midrule
\multirow{2}{*}{\makecell[l]{Gaming\\World Model}}
& Matrix-Game 2.0 & 0.216 & 0.222 & 0.326 & 0.255 \\
& LingBot-World & 0.400 & 0.383 & 0.337 & 0.373 \\
\midrule
\multirow{2}{*}{\makecell[l]{Robotics\\World Model}}
& Cosmos-Predict-2.5 & 0.415 & 0.405 & 0.378 & 0.399 \\
& WoW & 0.262 & 0.245 & 0.244 & 0.250 \\
\midrule
\multirow{4}{*}{\makecell[l]{General\\World Model}}
& Rolling Forcing & 0.359 & 0.332 & 0.272 & 0.321 \\
& LongLive & 0.379 & 0.365 & 0.345 & 0.363 \\
& Yume-1.5 & 0.338 & 0.334 & 0.231 & 0.301 \\
& Hunyuan-WorldPlay & 0.426 & 0.412 & 0.436 & 0.424 \\
\bottomrule
\end{tabularx}
\end{table}

\begin{table}[!htbp]
\centering
\scriptsize
\caption{Model-level interaction submetrics.}
\label{tab:model-level-interaction-results}
\setlength{\tabcolsep}{3pt}
\renewcommand{\arraystretch}{1.10}
\begin{tabularx}{\linewidth}{p{2.2cm}>{\raggedright\arraybackslash}Xcccccccc}
\toprule
\textbf{Category} & \textbf{Model} & \textbf{Chunk} & \textbf{Trans.} & \textbf{Global} & \textbf{Long Range} & \textbf{Global Text} & \textbf{CLIP Raw} & \textbf{CLIP Aux.} & \textbf{Interact.} \\
\midrule
\multirow{2}{*}{\makecell[l]{Gaming\\World Model}}
& Matrix-Game 2.0 & 0.123 & 0.109 & 0.058 & 0.058 & 0.057 & 0.237 & 0.185 & 0.113 \\
& LingBot-World & 0.709 & 0.751 & 0.864 & 0.875 & 0.850 & 0.314 & 0.570 & 0.734 \\
\midrule
\multirow{2}{*}{\makecell[l]{Robotics\\World Model}}
& Cosmos-Predict-2.5 & 0.704 & 0.705 & 0.791 & 0.807 & 0.775 & 0.313 & 0.565 & 0.707 \\
& WoW & 0.339 & 0.359 & 0.352 & 0.362 & 0.341 & 0.267 & 0.335 & 0.345 \\
\midrule
\multirow{4}{*}{\makecell[l]{General\\World Model}}
& Rolling Forcing & 0.600 & 0.666 & 0.671 & 0.699 & 0.641 & 0.327 & 0.635 & 0.636 \\
& LongLive & 0.552 & 0.398 & 0.613 & 0.636 & 0.585 & 0.323 & 0.615 & 0.526 \\
& Yume-1.5 & 0.590 & 0.745 & 0.697 & 0.726 & 0.667 & 0.306 & 0.530 & 0.649 \\
& Hunyuan-WorldPlay & 0.320 & 0.294 & 0.247 & 0.259 & 0.235 & 0.290 & 0.450 & 0.316 \\
\bottomrule
\end{tabularx}
\end{table}

\FloatBarrier

\section{Case Study}
\label{app:case-study}

We provide representative qualitative cases that illustrate how
WorldOlympiad diagnoses different failure modes beyond generic video quality.
Each case uses the same source prompt or reference context across models, so
the comparison focuses on model behavior rather than prompt variation.

\begin{table}[!htbp]
\centering
\small
\caption{Representative case studies and the corresponding diagnostic signals.}
\label{tab:case-study}
\setlength{\tabcolsep}{4pt}
\renewcommand{\arraystretch}{1.18}
\begin{tabularx}{\linewidth}{p{2.2cm}p{2.6cm}XX}
\toprule
\textbf{Case} & \textbf{Evaluation focus} & \textbf{Typical success pattern} & \textbf{Typical failure pattern} \\
\midrule
Physical dynamics & Gravity, impact, deformation, or phase transition & The object follows the expected temporal order, preserves contact constraints, and changes state gradually when required. & The object floats, teleports, deforms without contact, changes phase instantaneously, or violates the expected direction of motion. \\
3D consistency & Gaussian-splat reconstruction and camera trajectory & The reconstructed scene remains stable under novel views, with consistent foreground objects and plausible camera motion. & The reconstruction contains stretched geometry, missing background structure, unstable object identity, or camera motion that disagrees with the reference trajectory. \\
Interactive rollout & Chunk-level instruction following and transition coherence & Each generated chunk follows its action caption, and the next chunk preserves scene state, agent pose, and object layout. & The model resets the scene at chunk boundaries, ignores control changes, changes object identity, or accumulates visual drift over long horizons. \\
\bottomrule
\end{tabularx}
\end{table}

\begin{figure}[H]
\centering
\includegraphics[width=\linewidth]{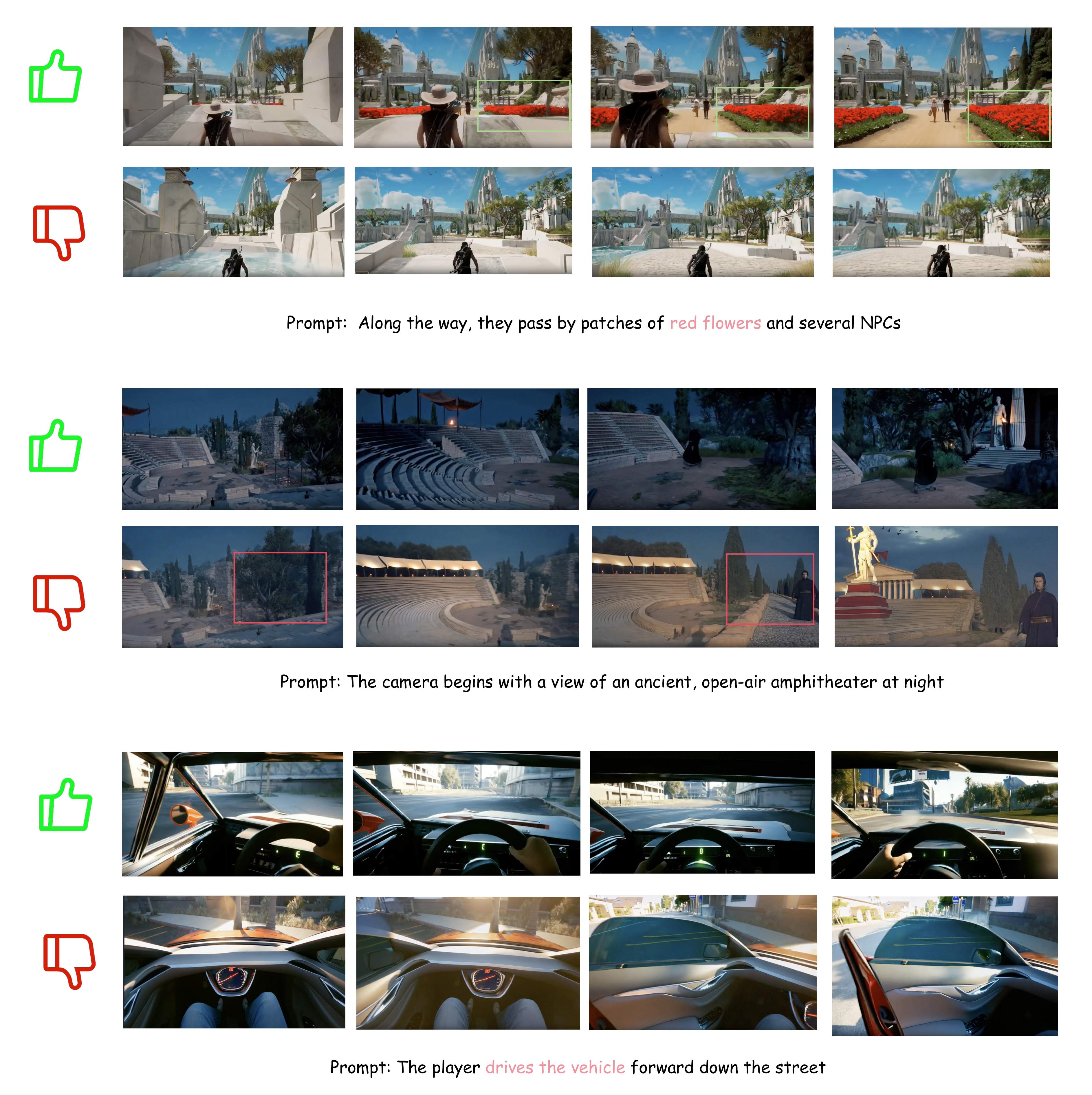}
\caption{
    Gaming case study from WorldOlympiad. The example highlights how interactive
    game rollouts expose action-following, scene-state preservation, and
    cross-chunk transition failures.
}
\label{fig:gaming-case-study}
\end{figure}

\paragraph{\textbf{Gaming case study.}}
Figure~\ref{fig:gaming-case-study} shows a gaming case study, where the main
diagnostic signals come from geometry consistency and interaction fidelity. The
geometry metric examines whether the generated video preserves a stable and
spatially coherent game scene under camera movement. In particular, it checks
whether the visual content remains consistent with the textual description of
the scene, including the expected environment, objects, style, and spatial
layout. When the camera moves, a strong model should maintain stable geometry
and avoid sudden scene deformation, object disappearance, or inconsistent
background structure. The interaction metric further evaluates whether the
generated rollout follows the intended action sequence and preserves the game
state across chunks. Failure cases include drifting away from the described
scene, producing unstable camera transitions, resetting the environment between
chunks, or generating actions that no longer match the corresponding captions.

\begin{figure}[H]
\centering
\includegraphics[width=\linewidth]{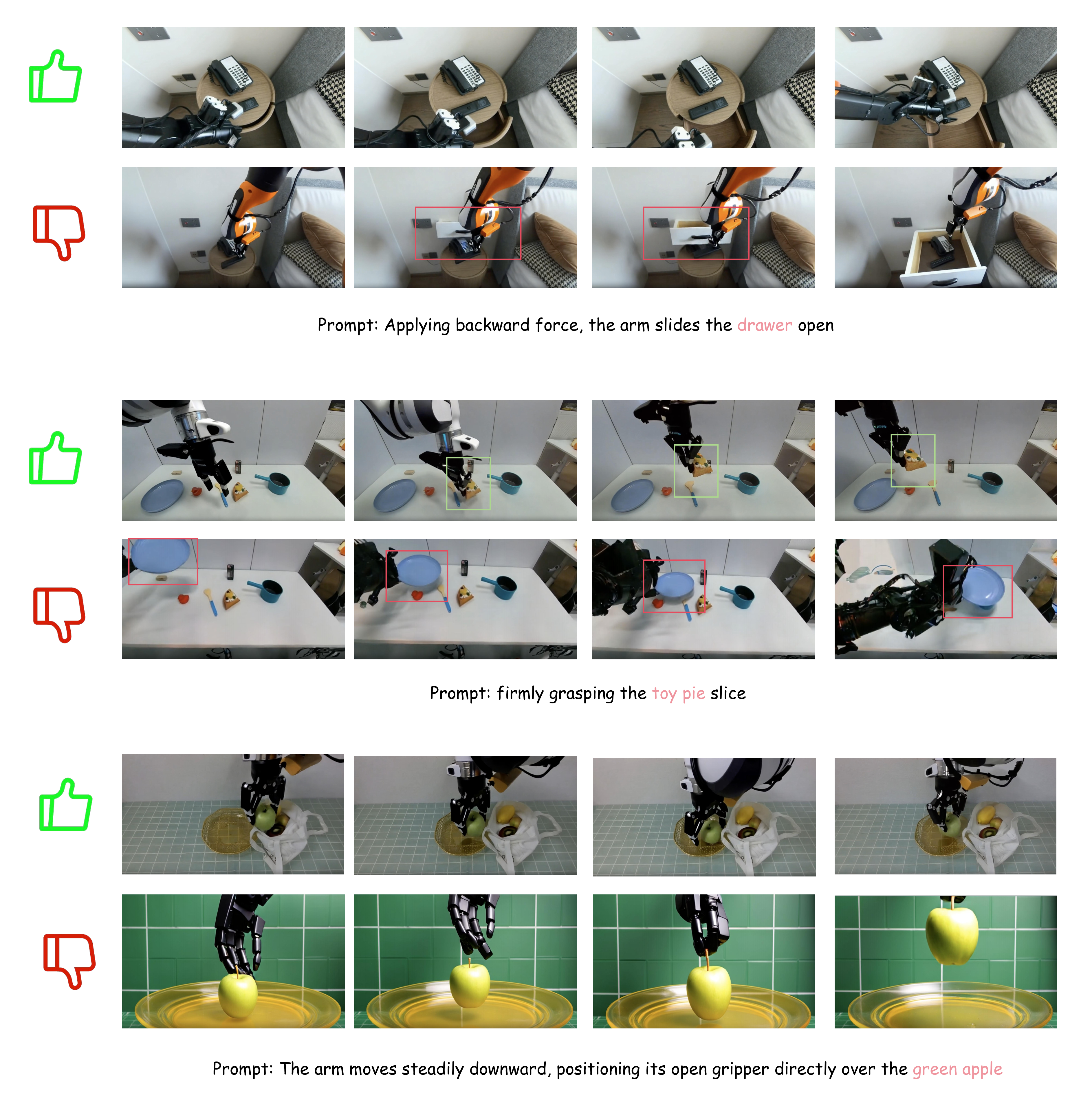}
\caption{
    Robotics case study from WorldOlympiad. The example visualizes how the
    benchmark diagnoses physical interaction, object-state consistency, and
    temporal coherence in robotics world-model rollouts.
}
\label{fig:embodied-case-study}
\end{figure}

\paragraph{\textbf{Robotics case study.}}
Figure~\ref{fig:embodied-case-study} presents an robotics manipulation case,
where WorldOlympiad jointly examines physical plausibility, scene-level
geometric consistency, and instruction-following behavior. For physical
evaluation, the case highlights failures such as an apple floating in mid-air
despite the absence of visible support, indicating a violation of gravity and
object-support constraints. For geometry evaluation, the benchmark further
checks whether the scene layout remains coherent throughout the rollout. For
example, a drawer may suddenly appear or disappear across frames, revealing
inconsistent spatial structure and unstable background reconstruction. For
interaction evaluation, the judge focuses on whether the robot follows the
intended manipulation instruction, such as reaching toward the correct object,
grasping the target item rather than a distractor, and maintaining a plausible
object state after contact. This case shows that visually plausible robotics
videos can still fail when object dynamics, scene consistency, or robot-action
alignment are not faithfully preserved.

\begin{figure}[H]
\centering
\includegraphics[width=\linewidth]{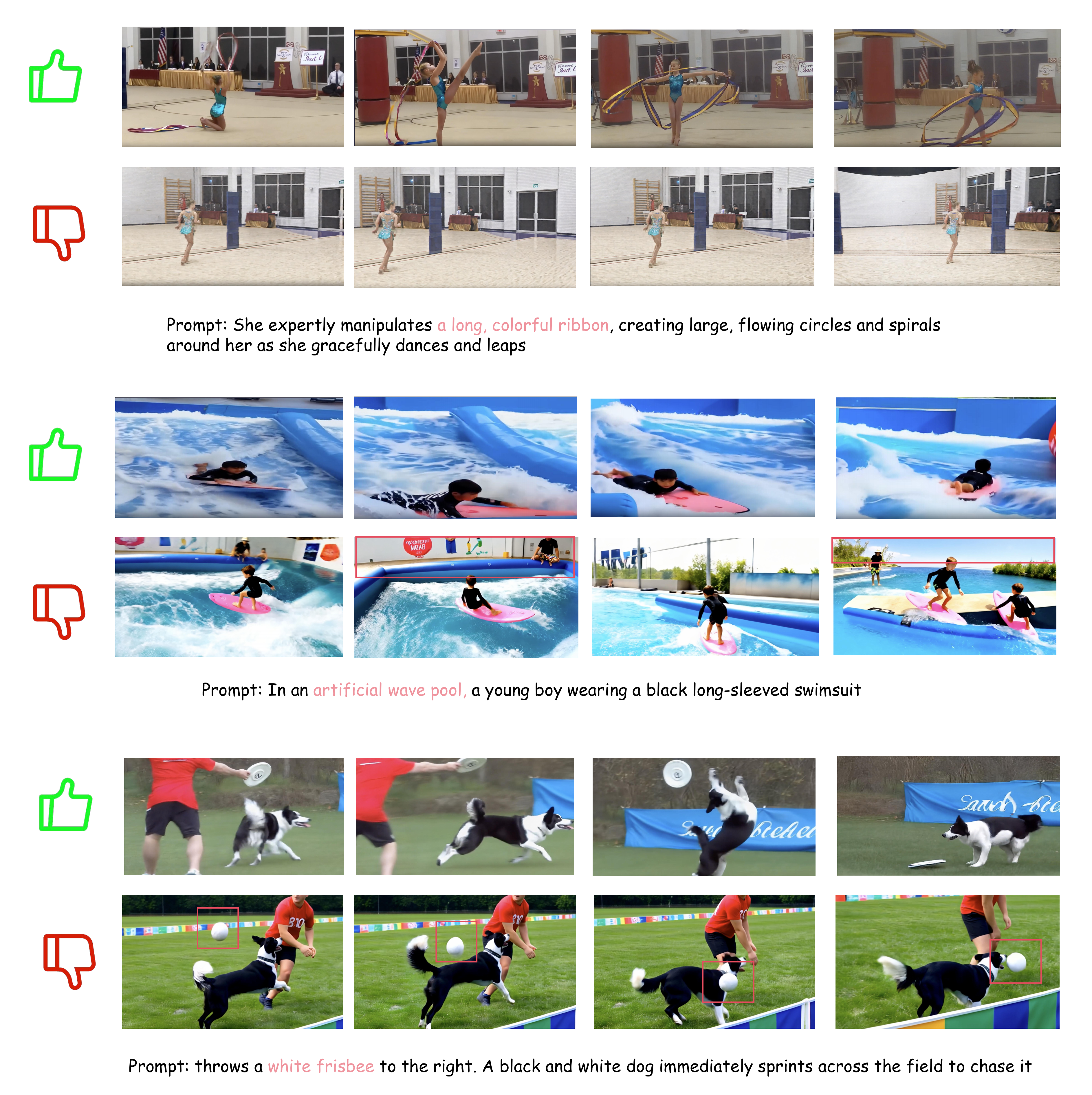}
\caption{
    Real-world case study from WorldOlympiad. The example illustrates how
    open-domain videos reveal geometric consistency, camera-motion, and
    long-range visual-coherence issues.
}
\label{fig:general-case-study}
\end{figure}

\paragraph{\textbf{General case study.}}
Figure~\ref{fig:general-case-study} presents a real-world case study, where all
three evaluation dimensions are informative. For physical evaluation, the case
checks whether the motion of a thrown frisbee follows a plausible trajectory,
rather than floating, stopping unnaturally, or changing direction without
visible cause. For geometry evaluation, the benchmark inspects whether the
scene remains spatially and semantically consistent over time. For instance, a
failure case may abruptly change an indoor scene into an outdoor scene,
indicating severe scene-level inconsistency and poor long-range coherence. For
interaction evaluation, the judge examines whether the generated video contains
meaningful temporal evolution rather than becoming overly static. A strong
sample should preserve realistic motion, maintain a coherent scene layout, and
continue to reflect the intended event throughout the video. These qualitative
examples demonstrate that WorldOlympiad can reveal complementary failure modes
in physical dynamics, 3D consistency, and interactive temporal behavior.

\section{Human Preference Study Details}
\label{app:human-preference-details}

The human preference alignment study in Table~\ref{tab:human-alignment} uses
the following annotation and aggregation protocol.

\paragraph{\textbf{Annotation protocol.}}
For each selected evaluation prompt, annotators compare anonymized generated
videos from the evaluated models under the same prompt or reference context.
Five annotators participate in the study. We sample 20 prompts from the
evaluation set and compare all \(\binom{8}{2}=28\) unordered model pairs under
each prompt, resulting in 560 prompt-level pairwise comparisons. Each comparison
is independently labeled by all five annotators, yielding 2,800 individual
preference labels. Annotators are instructed to judge the overall preference using four criteria:
visual quality, physical plausibility, temporal coherence, and interaction
fidelity. Model names are hidden during annotation. Ties are allowed when two
videos are indistinguishable or when their strengths and weaknesses are
balanced.

\begin{figure}[!htbp]
\centering
\includegraphics[width=\linewidth]{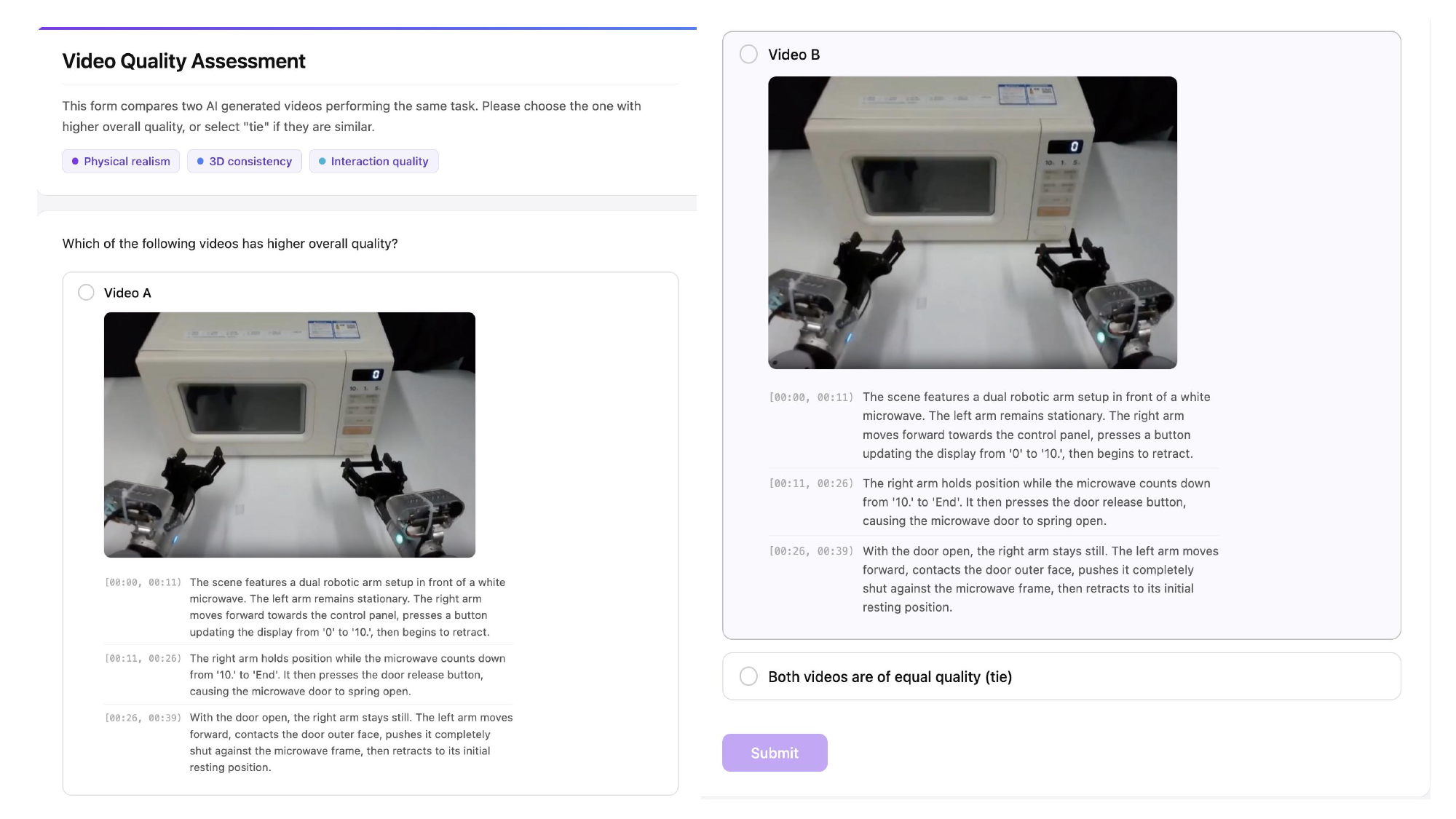}
\caption{Human preference annotation interface used in the alignment study.}
\label{fig:human-label-interface}
\end{figure}

\begin{tcolorbox}[promptbox,title={Human Preference Annotation Prompt}]
\small
\textbf{Input:} Two anonymized generated videos, Video A and Video B, produced
from the same prompt or reference context.\\
\textbf{Instruction:} Choose the video that better satisfies the prompt and
shows more realistic world dynamics. Consider visual quality, physical
plausibility, temporal coherence, object and scene consistency, and interaction
fidelity. If neither video is clearly better, select Tie.\\
\textbf{Output format:} Return one label from \{\texttt{A}, \texttt{B},
\texttt{Tie}\} and a one-sentence rationale.
\end{tcolorbox}

\paragraph{\textbf{Score aggregation.}}
Let \(y_{m,n,a}\) denote the preference outcome assigned by annotator \(a\) for
model \(m\) in comparison \(n\). A win contributes \(1\), a tie contributes
\(0.5\), and a loss contributes \(0\). We first average the five annotator
labels for each pairwise comparison and then compute the model-level preference
rate:
\[
\bar{y}_{m,n} = \frac{1}{5}\sum_{a=1}^{5} y_{m,n,a},\quad
S^{\mathrm{human}}_m = \frac{1}{N_m}\sum_{n=1}^{N_m} \bar{y}_{m,n},
\]
where \(N_m=140\) is the number of aggregated valid comparisons involving each
model.
Human ranks are obtained by sorting \(S^{\mathrm{human}}\) in descending order.
WorldOlympiad ranks are obtained by sorting the automatic overall evaluation
score in descending order, where \(S^{\mathrm{auto}}\) is the same
three-track average \(S_{\mathrm{all}}\) used in the main benchmark table.

\paragraph{Rank correlation.}
We measure alignment using Spearman's rank correlation:
\[
\rho = 1 - \frac{6\sum_{m=1}^{M} d_m^2}{M(M^2-1)},
\quad
d_m = r^{\mathrm{human}}_m - r^{\mathrm{auto}}_m,
\]
where \(M\) is the number of evaluated models. For the eight models with human
preference annotations, the resulting correlation is \(0.95\), indicating strong
agreement between human preference and the WorldOlympiad automatic ranking.

The rank disagreements occur only in two adjacent pairs: LongLive and Yume-1.5,
and Matrix-Game 2.0 and WoW. These swaps have a limited effect on the overall
correlation and suggest that the automatic evaluator preserves the main model
ordering while still exposing borderline cases where human preference and
rubric-based automatic scores differ.


\end{document}